\documentclass[letterpaper]{article} 
\usepackage{aaai24}  
\usepackage{times}  
\usepackage{helvet}  
\usepackage{courier}  
\usepackage[hyphens]{url}  
\usepackage{graphicx} 
\usepackage{natbib}  
\usepackage{caption} 
\usepackage{algorithm}
\usepackage{algorithmic}
\usepackage{amsmath}
\usepackage{amsthm}
\usepackage{amssymb}
\usepackage{multicol}
\usepackage{xcolor}
\usepackage{newfloat}
\usepackage{listings}
\DeclareCaptionStyle{ruled}{labelfont=normalfont,labelsep=colon,strut=off} 
\floatstyle{ruled}
\newfloat{listing}{tb}{lst}{}
\floatname{listing}{Listing}
\newtheorem{theorem}{Theorem}[section]
\newtheorem{proposition}[theorem]{Proposition}
\newtheorem{lemma}[theorem]{Lemma}
\newtheorem{corollary}[theorem]{Corollary}

\newtheorem{assumption}[theorem]{Assumption}

\title{Hierarchical Topology Isomorphism Expertise Embedded\\Graph Contrastive Learning}
\author{
    Jiangmeng Li \textsuperscript{\rm 1 \rm 2}\equalcontrib,
    Yifan Jin\textsuperscript{\rm 1 \rm 3}\equalcontrib,
    Hang Gao\textsuperscript{\rm 1},
    Wenwen Qiang\textsuperscript{\rm 1}\thanks{Corresponding author.},
    Changwen Zheng\textsuperscript{\rm 1 \rm 3}$^\dagger$,
    Fuchun Sun\textsuperscript{\rm 1 \rm 4}
}
\affiliations{
    \textsuperscript{\rm 1}Science \& Technology on Integrated Information System Laboratory, Institute of Software Chinese Academy of Sciences\\

    \textsuperscript{\rm 2}State Key Laboratory of Intelligent Game\\
    \textsuperscript{\rm 3}University of Chinese Academy of Sciences,
    \textsuperscript{\rm 4}Tsinghua University\\
    \{jiangmeng2019,yifan2020,gaohang,qiangwenwen,changwen\}@iscas.ac.cn, fcsun@mail.tsinghua.edu.cn
}

\usepackage{bibentry}

\begin{document}

\maketitle

\begin{abstract}
Graph contrastive learning (GCL) aims to align the positive features while differentiating the negative features in the latent space by minimizing a pair-wise contrastive loss. As the embodiment of an outstanding discriminative unsupervised graph representation learning approach, GCL achieves impressive successes in various graph benchmarks. However, such an approach falls short of recognizing the topology isomorphism of graphs, resulting in that graphs with relatively homogeneous node features cannot be sufficiently discriminated. By revisiting classic graph topology recognition works, we disclose that the corresponding expertise intuitively complements GCL methods. To this end, we propose a novel hierarchical topology isomorphism expertise embedded graph contrastive learning, which introduces knowledge distillations to empower GCL models to learn the hierarchical topology isomorphism expertise, including the graph-tier and subgraph-tier. On top of this, the proposed method holds the feature of plug-and-play, and we empirically demonstrate that the proposed method is universal to multiple state-of-the-art GCL models. The solid theoretical analyses are further provided to prove that compared with conventional GCL methods, our method acquires the tighter upper bound of Bayes classification error. We conduct extensive experiments on real-world benchmarks to exhibit the performance superiority of our method over candidate GCL methods, e.g., for the real-world graph representation learning experiments, the proposed method beats the state-of-the-art method by 0.23\% on unsupervised representation learning setting, 0.43\% on transfer learning setting. 
Our code is available at https://github.com/jyf123/HTML.

\end{abstract}

\section{Introduction} \label{sec:intro}
Unsupervised graph representation learning aims to capture the discriminative information from graphs without manual annotations. The intuition behind such a learning approach is that the covariate shift problem \cite{heckman1979sample, shimodaira2000improving} occurs when the data distributions in the source labeled and target unlabeled domains are different. Thus, the \textit{identically distributed} assumption is violated, halting the forthright application of the model trained in the supervised manner on the target unlabeled dataset. In view of that, the contrastive learning paradigm has achieved impressive successes in a spectrum of practical fields, such as computer vision \cite{DBLP:conf/cvpr/WuXYL18,DBLP:conf/icml/ChenK0H20,DBLP:conf/icml/QiangLZ0X22}, natural language processing \cite{DBLP:conf/emnlp/GaoYC21,DBLP:conf/iclr/QuSSSC021}, signal processing \cite{DBLP:conf/www/YaoZZZZZH22,DBLP:conf/icassp/TangZWCX22}, etc. As tractable surrogates, the contrastive approaches further establish promising capacity in the field of unsupervised graph learning, namely GCL methods \cite{DBLP:conf/nips/YouCSCWS20}. The primary principle of GCL is jointly differentiating features of heterogeneous samples while gathering features of homogeneous samples.

A major challenge for state-of-the-art (SOTA) GCL methods is sufficiently exploring \textit{heterogeneous} discriminative information. Orthogonal to the natural data structures, e.g., images and videos, as the artificially derived discrete data structure, graphs contain two critical ingredients of potential discriminative information: feature-level information and topology-level information. Benefiting from the powerful non-linear mapping capability of graph neural networks (GNNs) \cite{DBLP:journals/corr/KipfW16,DBLP:journals/corr/abs-1710-10903,gao2023robust}, GCL methods can practically explore feature-level discriminative information. However, recent works \cite{DBLP:journals/corr/abs-1810-00826,DBLP:conf/iclr/Wijesinghe022} demonstrate that compared with topology-expertise-based approaches, GNN-based GCL methods can only model relatively limited topology-level discriminative information, e.g., \cite{DBLP:journals/corr/abs-1810-00826} empirically substantiate that specific GNN-based methods \cite{DBLP:journals/corr/KipfW16,DBLP:conf/nips/HamiltonYL17} fall short of discriminating the topology structures of graphs that the Weisfeiler-Lehman-based approach can explicitly discriminate.
\begin{figure*}
  \centering
  \includegraphics[scale=0.4]{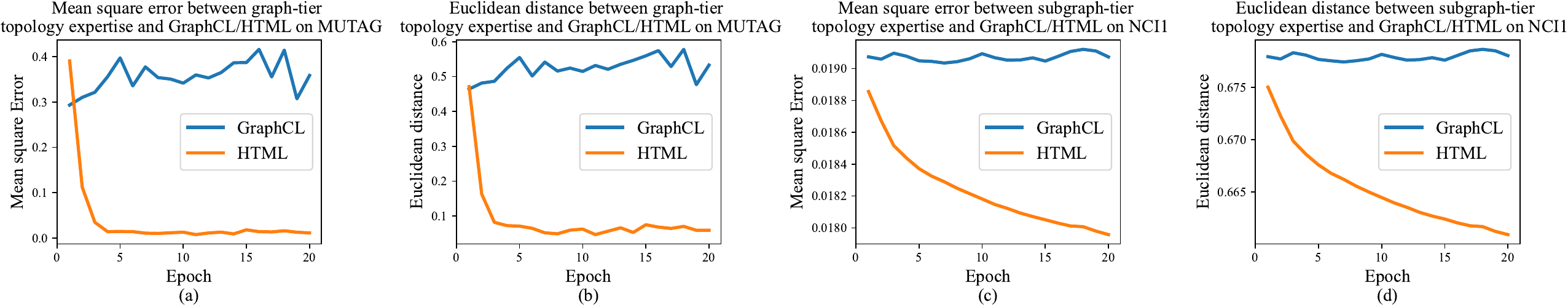}
  \caption{Motivating experiments to explore whether canonical GCL methods can learn the topology expertise on graph benchmarks. For the fair comparison principle, we adopt two metrics to measure the error or similarity between the candidate methods and the topology expertise, containing the graph-tier and subgraph-tier, during training. The volatile result curves show that canonical GCL methods, e.g., GraphCL, can barely learn topology expertise. Yet, the consistent result curves in the downtrend substantiate that the proposed method, i.e., HTML, can indeed introduce the topology expertise in GCL methods.}
\label{motivation}

\end{figure*}
To this end, we are primarily dedicated to exploring the difference in knowledge between the topology-expertise-based approaches and GNN-based GCL methods with respect to capturing the discriminative information. As demonstrated in Figure \ref{motivation}, the experimental exploration attests that canonical GNN-based GCL methods \textit{cannot} effectively learn the topology expertise, and accordingly, introducing such graph expertise into the GNN-based GCL method incurs a precipitous rise in the discriminative performance. Concretely, we empirically derive a conclusion: the topology expertise is complementary to GNN-based GCL methods, and prudently aggregating the topology-level and feature-level discriminative information can explicitly improve the model's performance on unsupervised graph prediction.

Considering the aforementioned conclusion, we explore implicitly introducing the graph topology expertise into GCL models and thus propose a plug-and-play method, namely \textit{hierarchical topology isomorphism expertise embedded graph contrastive learning} (HTML), which leverages the knowledge distillation technique to prompt GCL models to learn the topology expertise. Specifically, by revisiting the graph topology expertise, we disclose that the principal objective of such expertise is determining the \textit{isomorphism} of candidate graph units, and the determination of whether candidate graphs are graph-tier isomorphic is identified as a \textit{non-deterministic polynomial immediate problem} (NPI problem) \cite{DBLP:journals/corr/Babai15}. Yet, state-of-the-art graph isomorphism studies demonstrate that there exist certain closed-form analytical solutions in polynomial time, e.g., Weisfeiler-Lehman (WL) test \cite{weisfeiler1968reduction}, which can fit most cases in the graph isomorphism problem, such that introducing the guidance of graph topology isomorphism expertise into GCL methods is theoretically feasible. Considering there are still certain cases that cannot be solved by the deterministic methods in polynomial-time, we propose to introduce the graph isomorphism expertise in an implicit instead of explicit manner. Accordingly, the topology isomorphism expertise adheres to the hierarchical principle and can be further divided into the graph-tier and the subgraph-tier. We elaborate that the graph-tier topology isomorphism expertise dedicates to globally classifying graphs with different topological structures, while the subgraph-tier topology isomorphism expertise focuses on discriminating graphs regarding certain critical subgraphs, e.g., functional groups for a chemical molecule, by leveraging the weighting strategy. Concretely, in view of the conducted exploration, we reckon that HTML can generally improve GCL methods with confidence. The thoroughly proved theoretical analyses are provided to demonstrate that HTML can tighten the upper bound of Bayes classification error acquired by conventional GCL methods. The extensive experiments on benchmarks demonstrate that the proposed method is simple yet effective. In detail, HTML generally exhibits performance superiority over candidate GCL methods under the unsupervised learning experimental setting. As a plug-and-play module, HTML consistently promotes the performance of baselines by significant margins.
\textbf{Contributions}:

\begin{itemize}

    \item We disclose the differences between the topology isomorphism expertise approach and the GNN-based GCL method and further introduce an empirical exploration, which demonstrates that prudently aggregating the topology isomorphism expertise and the knowledge contained by canonical GCL methods can effectively improve the model performance.

    \item We propose the novel HTML, orthogonal to existing methods, to introduce the hierarchical topology isomorphism expertise learning into the GCL paradigm, thereby promoting the model to capture discriminative information from graphs in a plug-and-play manner.

    \item The performance rises of GCL methods incurred by leveraging the proposed HTML are theoretically supported since HTML can tighten the Bayes error upper bound of canonical GCL methods on graph learning benchmarks.

    \item Empirically, experimental evaluations on various real-world datasets demonstrate that HTML yields wide performance boosts on conventional GCL methods.
    
\end{itemize}

\section{Related Works} \label{sec:related_works}
 Graph neural networks (GNNs) are a group of neural networks which can effectively encode graph-structured data. Since the inception of the first GNNs model, multiple variants have been proposed, including GCN \cite{DBLP:journals/corr/KipfW16}, GAT \cite{DBLP:journals/corr/abs-1710-10903}, GraphSAGE \cite{DBLP:conf/nips/HamiltonYL17}, GIN \cite{DBLP:journals/corr/abs-1810-00826}, and others. These models learn discriminative graph representations guided by the labels of graph data. Since the annotation of graph data, such as the categories of biochemical molecules, requires the support of expert knowledge, it is not easy to obtain large-scale annotated graph datasets \cite{DBLP:conf/nips/YouCSCWS20}. This leads to the limitation of supervised graph representation learning.

As one of the most effective self-supervised methods, contrastive learning (CL) aims to learn discriminative representations from unlabelled data \cite{li2022metamask}. With the principle of closing positive and moving away from negative pairs, CL methods, such as SimCLR \cite{DBLP:conf/icml/ChenK0H20}, MoCo \cite{DBLP:conf/cvpr/He0WXG20}, BYOL \cite{DBLP:conf/nips/GrillSATRBDPGAP20}, MetAug \cite{li2022metaug}, and Barlow Twins \cite{DBLP:conf/icml/ZbontarJMLD21}, have achieved outstanding success in the field of computer vision \cite{DBLP:conf/cvpr/WuXYL18,DBLP:conf/icml/ChenK0H20,DBLP:conf/icml/QiangLZ0X22}.

Benefiting from GNNs and CL methods, GCL is proposed \cite{DBLP:conf/iclr/VelickovicFHLBH19,DBLP:conf/iclr/SunHV020,DBLP:conf/icml/HassaniA20,DBLP:conf/nips/YouCSCWS20}. Due to the difference between graph and image, the GCL methods focus more on constructing graph augmentation. Specifically, GraphCL \cite{DBLP:conf/nips/YouCSCWS20} first introduces the perturbation invariance for GCL and proposes four types of graph augmentation: node dropping, edge perturbation, attribute masking, and subgraph. Noting that using a complete graph leads to high time and space utilization, Subg-Con \cite{DBLP:conf/icdm/JiaoXZ0ZZ20} proposes to sample subgraphs to capture structural information. To enrich the semantic information in the sampled subgraphs, MICRO-Graph \cite{DBLP:journals/corr/abs-2012-12533} proposes to generate informative subgraphs by learning graph motifs. Besides, the choice of graph augmentation requires rules of thumb or trial-and-error, resulting in time-consuming and labor-intensive. Thus, JOAO \cite{DBLP:conf/icml/YouCSW21} introduces a bi-level optimization framework that automatically selects data augmentations for particular graph data. RGCL \cite{DBLP:conf/icml/LiWZW0C22} argues that randomly destroying the properties of graphs will result in the loss of important semantic information in the augmented graph and proposes a rationale-aware graph augmentation method. For graph invariance in graph augmentation, SPAN \cite{DBLP:conf/iclr/LinCW23} finds that previous studies have only performed topology augmentation in the spatial domain, therefore introducing a spectral perspective to guide topology augmentation. Noting that existing GCL algorithms ignore the intrinsic hierarchical structure of the graph, HGCL \cite{DBLP:journals/nn/JuGLWYZZ23} proposes Hierarchical GCL consisting of three components: node-level CL, graph-level CL, and mutual CL. Despite its attention to the neglect of the intrinsic hierarchical structure, compared to HTML, it still focuses only on the semantic information, i.e., the feature-level information. Another important issue with GCL is negative sampling. BGRL \cite{DBLP:conf/iclr/ThakoorTAADMVV22} proposes a method that uses simple graph augmentation and does not have to construct negative samples while effectively scaling to large-scale graphs. To address the issue of sampling bias in GCL, PGCL \cite{DBLP:journals/corr/abs-2106-09645} proposes a negative sampling method based on semantic clustering. In contrast to previous approaches, HTML introduces topology-level information into GCL, which has never been approached from a novel perspective, i.e., neither constructing graph augmentation nor improving negative sampling.

\section{Preliminary} \label{sec:preliminary}
GCL aims at maximizing the mutual information of positive pairs while minimizing that of negative pairs, which is capable of learning discriminative graph representations without annotations. Generally, GCL methods consist of three components, as shown in the middle of Figure \ref{framework}.\\
\textbf{Graph augmentation}. Let $G = (V, E)$ denotes a graph, $V$ is the node set, and $E$ is the edge set. Graph $G$ is first augmented into two different views $\hat{G}^i$ and $\hat{G}^j$ by node dropping, edge perturbation \cite{DBLP:conf/nips/YouCSCWS20}, rationale sampling \cite{DBLP:conf/icml/LiWZW0C22}, etc. \\
\textbf{Graph encoder}. The graph encoder, e.g., specific GNNs with shared parameters, is used to extract graph representations $Z^i$ and $Z^j$ for views $\hat{G}^i$ and $\hat{G}^j$. \\
\textbf{Contrastive loss function}. Before computing the contrastive loss, a projection head is commonly applied to map $Z^i$ and $Z^j$ into $\tilde{Z}^i$ and $\tilde{Z}^j$. Then, by considering two different views of the same graph, such as $\hat{G}^i$ and $\hat{G}^j$, as positive pairs while views of different graphs are negative pairs and adopting the normalized temperature-scaled cross-entropy loss (NT-Xent) \cite{DBLP:journals/corr/abs-1807-03748,DBLP:conf/cvpr/WuXYL18,DBLP:conf/nips/Sohn16}, the loss can be computed as follows:
\begin{equation}
    \mathcal{L}_{c} = \frac{1}{N} \sum_{n=1}^N log{\frac{-\textrm{exp}\left(\textrm{sim}\left(\tilde{Z}_{n,i},\tilde{Z}_{n,j}\right)/\tau\right)}{\sum_{n^{'}=1,n^{'}\neq n}^N \textrm{exp}\left(\textrm{sim}\left(\tilde{Z}_{n,i},\tilde{Z}_{n^{'},j}\right)/\tau\right)}},
    \label{4-6}
\end{equation}
where $\textrm{sim}(\cdot,\cdot)$ denotes cosine similarity \cite{DBLP:journals/isci/XiaZL15}, which can be calculated as $\textrm{sim}(\tilde{Z}_{n,i},\tilde{Z}_{n,j})=\tilde{Z}_{n,i} \tilde{Z}_{n,j}^\top/{\left\| \tilde{Z}_{n,i} \right\| \left\| \tilde{Z}_{n,j}^\top \right\|}$, $N$ is the number of graphs, and $\tau$ is the temperature coefficient.

\section{Methodology} \label{sec:method}
\begin{figure*}
  \centering
  \includegraphics[scale=0.42]{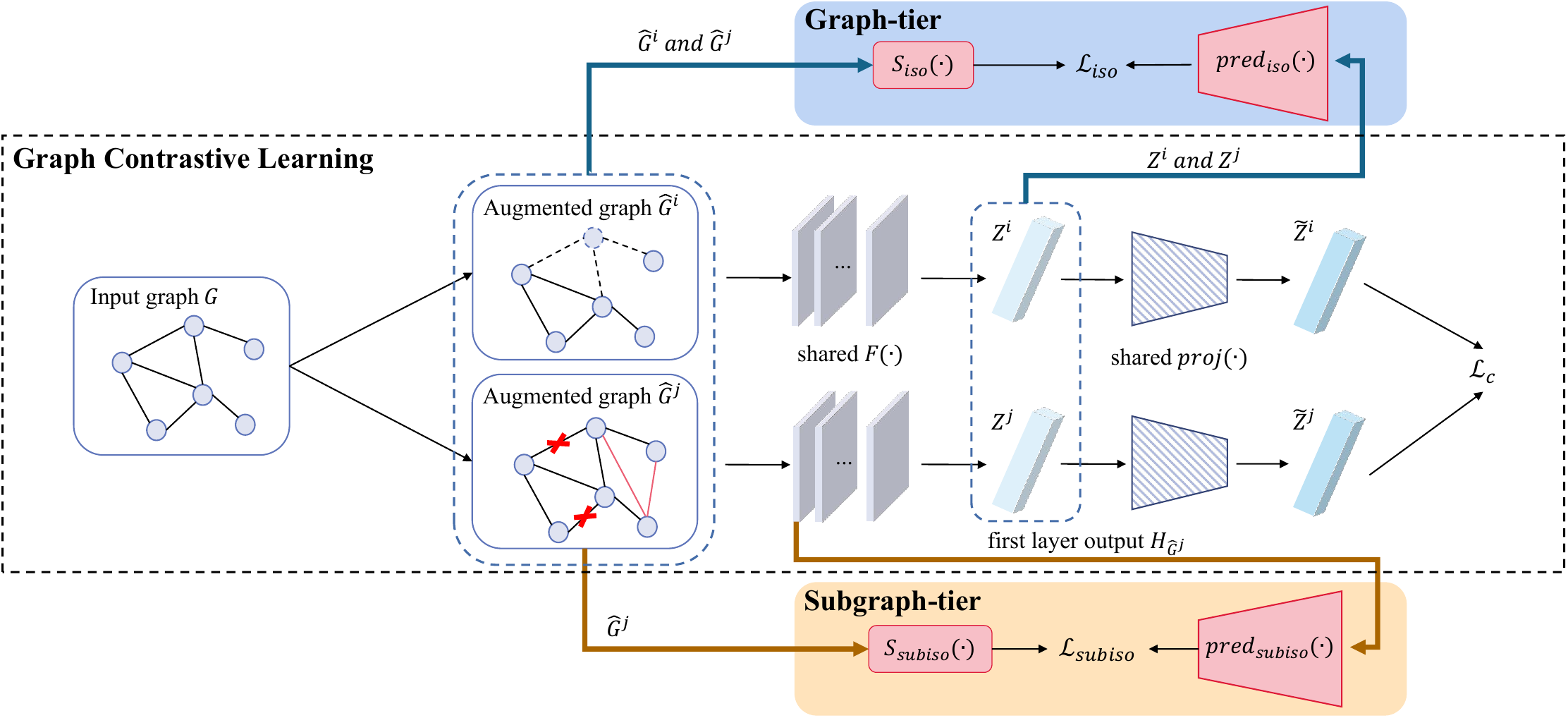}
  \caption{Architecture of the proposed HTML based on a canonical GCL approach.}
\label{framework}

\end{figure*}
To enable GNN-based GCL methods to model topology-level discriminative information, we implicitly incorporate topology isomorphism expertise into GCL models at both the graph-tier and the subgraph-tier. The overall framework is illustrated in Figure \ref{framework}.

\subsection{Learning Graph-tier Topology Isomorphism Expertise}
With the aim of empowering the graph encoder $F(\cdot)$ to learn graph-tier topology isomorphic expertise, we propose an expert system $\mathcal{S}_{iso}(\cdot)$ to measure the isomorphic similarity between two graphs. The graph-tier topology isomorphism measures the topological equivalence of two graphs. Determining whether two graphs are graph-tier topology isomorphic remains an NPI problem. Among many methods \cite{DBLP:journals/jacm/CorneilG70, DBLP:journals/jsc/McKayP14, DBLP:journals/corr/Babai15, weisfeiler1968reduction} for this problem, the WL test \cite{weisfeiler1968reduction}, especially the 1-dimensional WL test, is a relatively effective method that can correctly deterministically confirm whether two graphs are graph-tier isomorphic in most cases \cite{DBLP:conf/focs/BabaiK79}. Specifically, given a graph with pre-defined node labels, the 1-dimensional WL test iteratively 1) aggregates the labels of neighbors to the original node labels, 2) updates node labels by the label compression and relabeling, and eventually obtains the set of node labels after several iterations. Accordingly, we build the proposed learning procedure of graph-tier topology isomorphism expertise. Given two augmented graphs $\hat{G}^i$ and $\hat{G}^j$ of graph $G$ sampled \textit{i.i.d} from a dataset with $N$ graphs, the expert system $\mathcal{S}_{iso}(\cdot)$ first performs the 1-dimensional WL test to obtain the node label sets $A^i$ and $A^j$, respectively. In practice, the number of iterations of the 1-dimensional WL test is the same as the number of layers of the encoder $F(\cdot)$. The graph-tier isomorphic similarity between the graphs can be computed by adopting the Jaccard Coefficient \cite{jaccard1912distribution}:
\begin{equation}
    y_{iso} = \frac{\left|A^i\cap A^j\right|}{\left|A^i \cup A^j\right|},
    \label{4-1}
\end{equation}
where $y_{iso}$ denotes the graph-tier topology isomorphic similarity between $\hat{G}^i$ and $\hat{G}^j$.

To introduce the graph-tier topology isomorphism expertise to GCL methods, we propose a self-supervised distillation task. Specifically, we train $F(\cdot)$ to predict the isomorphic similarity of $\hat{G}^i$ and $\hat{G}^j$ based on the graph representation obtained by $F(\cdot)$. For the prediction, the representations of $\hat{G}^i$ and $\hat{G}^j$ obtained by $F(\cdot)$ are concatenated and fed into a multi-layer perceptron (MLP), which can be treated as a prediction head. This process can be formulated by
\begin{equation}
    \hat{y}_{iso} = \textrm{MLP} \left([Z^i;Z^j]\right).
    \label{4-2}
\end{equation}

We adopt the mean square error \cite{allen1971mean} to measure the ability of $F(\cdot)$ to learn the topology isomorphism expertise in the graph-tier and promote $F(\cdot)$ to capture the topology discriminative information:
\begin{equation}
    \mathcal{L}_{iso} = \frac{1}{N} \sum_{n=1}^N \left(y_{iso}^n-\hat{y}_{iso}^n\right)^2.
    \label{4-3}
\end{equation}

\subsection{Learning Subgraph-tier Topology Isomorphism Expertise}

Graph-tier topology isomorphism expertise treats one graph as a whole, while in practice, specific subgraphs of a graph provide relatively significant contributions in the discrimination of graphs on certain downstream tasks. For instance, the functional groups of a chemical molecule have a decisive influence on certain properties. Therefore, we argue that learning the subgraph-tier topology isomorphism expertise can further promote the discriminative performance of GCL methods. Inspired by \cite{DBLP:conf/iclr/Wijesinghe022}, we introduce an expert system $\mathcal{S}_{subiso}(\cdot)$ to guide $F(\cdot)$ to explore the subgraph-tier topology isomorphism expertise. To capture the fine-grained topology information in a graph, \textit{structural coefficients} \cite{DBLP:conf/iclr/Wijesinghe022} are introduced. By modeling the structural characteristics of the \textit{overlap subgraph}\footnote{Refer to \textbf{Appendix \ref{app:example}} for examples of the overlap subgraph.} between adjacent nodes, the structural coefficients can satisfy the properties of local closeness, local denseness, and isomorphic invariance. Specifically, local closeness presents that the structural coefficients are positively correlated with the number of the overlap subgraph nodes; local denseness presents the positive correlation between the structural coefficients and the number of the overlap subgraph edges; isomorphic invariant presents that the structural coefficients are identical when the overlap subgraphs are isomorphic. Given a node $v$ in a graph $G$ sampled \textit{i.i.d} from a dataset with $N$ graphs, the \textit{neighborhood subgraph} of node $v$, denoted as $O_v$, is composed by itself, the nodes directly connected to $v$, and the corresponding edges. For two adjacent nodes $v$ and $u$, the overlap subgraph between them is defined as the intersection of their neighborhood subgraphs, i.e., $O_{vu} = O_v \cap O_u$. Formally, the structural coefficient $w_{vu}$ of $v$ and $u$ can be implemented by
\begin{equation}
    w_{vu} = \frac{|N^{E_{vu}}|}{|N^{V_{vu}}|\cdot|N^{V_{vu}}-1|}|N^{V_{vu}}|^\lambda,
    \label{a1}
\end{equation}
where $N^{V_{vu}}$ and $N^{E_{vu}}$ denote the aggregated numbers of nodes and edges of $O_{vu}$, respectively, and $\lambda\textgreater0$ is a hyperparameter. Then, $w_{vu}$ is normalized by $\tilde{w}_{vu} = \frac{w_{vu}}{\sum_{u \in \mathcal{N}(v)}w_{vu}}$, where $\mathcal{N}(v)$ is the neighbor set of $v$. For guiding the encoder $F(\cdot)$ to learn the fine-grained topology expertise, the structural coefficient matrix of $G$ is derived to constitute the subgraph-tier topology isomorphism expertise $y_{subiso}$. 

We further leverage the node representation $H_G$ obtained from the first layer of $F(\cdot)$ to fit the structural coefficient matrix:
\begin{equation}
    \hat{y}_{subiso} = \textrm{MLP}\left(\textrm{AUTOCOR}\left(\textrm{MLP}\left(H_G\right)\right)\right),
    \label{4-4}
\end{equation}
where $\hat{y}_{subiso}$ denotes the predicted subgraph-tier topology isomorphism, $\textrm{MLP}$s can be synthetically treated as the projection head, and $\textrm{AUTOCOR}\left(\cdot\right)$ denotes the autocorrelation matrix computation function \cite{berne1966calculation}. 

The mean square error loss is used to align the predictions of the expert system and $F(\cdot)$, thereby deriving
\begin{equation}
    \mathcal{L}_{subiso} = \frac{1}{N} \sum_{n=1}^N \left(y_{subiso}^n-\hat{y}_{subiso}^n\right)^2.
    \label{4-5}
\end{equation}

\subsection{Training Pipeline}

To jointly learn the graph-tier and subgraph-tier topology isomorphism expertise, we introduce $\mathcal{L}_{iso}$ and $\mathcal{L}_{subiso}$ into the GCL paradigm and derive the ultimate objective as follows:
\begin{equation}
    \mathcal{L} = \mathcal{L}_{c} + \alpha \mathcal{L}_{iso} + \beta \mathcal{L}_{subiso},
    \label{4-7}
\end{equation}
where $\alpha$ and $\beta$ are coefficients balancing the impacts of $\mathcal{L}_{iso}$ and $\mathcal{L}_{subiso}$ during training, respectively. The detailed training process is shown in Algorithm \ref{alg:html}.

\begin{algorithm}[ht]
	\vskip 0.in
 \caption{HTML}
 \label{alg:html}
	\begin{algorithmic}
		\STATE {\bfseries Input:} two augmented graphs $\hat{G}^i$ and $\hat{G}^j$ of graph $G$, graph encoder $F(\cdot)$, projection head $proj(\cdot)$, node representation $H_{\hat{G}^j}$ obtained by the first layer of $F(\cdot)$, temperature parameter $\tau$, and hyperparameters $\alpha$, $\beta$.\\
		\FOR{$t$-th training iteration}
            \STATE $Z^i = F(\hat{G}^i), Z^j = F(\hat{G}^j)$
            \STATE $\tilde{Z}^i = proj(Z^i), \tilde{Z}^j = proj(Z^j)$
		\STATE $\# \ graph \ contrastive \ learning \ loss$
            \STATE $\mathcal{L}_{c} = -\frac{1}{N} \sum_{n=1}^N \log{\frac{\textrm{exp}\left(\textrm{sim}\left(\tilde{Z}_{n,i},\tilde{Z}_{n,j}\right)/\tau\right)}{\sum_{n^{'}=1,n^{'}\neq n}^N \textrm{exp}\left(\textrm{sim}\left(\tilde{Z}_{n,i},\tilde{Z}_{n^{'},j}\right)/\tau\right)}}$ \\\
           
            \STATE $\# \ learning \ graph-tier$
            \STATE $\# \ topology \ isomorphism \ expertise$
		\STATE Execute expert system $\mathcal{S}_{iso}(\cdot)$ to obtain graph-tier topology isomorphism expertise $y_{iso}$ of $\hat{G}^i$ and $\hat{G}^j$. \\
            \STATE $\hat{y}_{iso} = \textrm{MLP} \left([Z^i;Z^j]\right)$ \\
            \STATE $\mathcal{L}_{iso} = \frac{1}{N} \sum_{n=1}^N \left(y_{iso}^n-\hat{y}_{iso}^n\right)^2$ \\\
            
            \STATE $\# \ learning \ subgraph-tier$
            \STATE $\# \ topology \ isomorphism \ expertise$
 		\STATE Execute expert system $\mathcal{S}_{subiso}(\cdot)$ to obtain subgraph-tier topology isomorphism expertise $y_{subiso}$ of $\hat{G}^i$. \\
            \STATE $\hat{y}_{subiso} = \textrm{MLP}\left(\textrm{AUTOCOR}\left(\textrm{MLP}\left(H_{\hat{G}^i}\right)\right)\right)$ \\
            \STATE $\mathcal{L}_{subiso} = \frac{1}{N} \sum_{n=1}^N \left(y_{subiso}^n-\hat{y}_{subiso}^n\right)^2$\\\
            \STATE $\# \ training \ loss$ \\
            \STATE $\mathcal{L} = \mathcal{L}_{c} + \alpha \mathcal{L}_{iso} + \beta \mathcal{L}_{subiso}$ \\
            \STATE Update encoder $F(\cdot)$ via $\mathcal{L}$.
        \ENDFOR
	\end{algorithmic}
	\vskip -0.in
	
\end{algorithm}

\section{Theoretical Analyses} \label{sec:theory}
Within the canonical self-supervised GCL paradigm, we demonstrate that compared with conventional graph contrastive methods, the proposed HTML achieves a relatively tighter Bayes error bound, thereby validating the performance superiority of HTML from the theoretical perspective. For conceptual simplicity, we demonstrate the Bayes error bound on a binary classification problem within a 1-dimensional latent space. First, we hold an assumption:
\begin{assumption}
    \label{ass:1}
    Considering the universality of Gaussian distributions in statistics, we assume that the prior distributions of candidate categories, i.e., $\mathcal{P}_{\mathcal{C}_1}$ and $\mathcal{P}_{\mathcal{C}_2}$, adhere to the central limit theorem \cite{rosenblatt1956central} on the target bi-classification task $\mathcal{T}^b$, such that $\left\{X_i\right\}_{i=0}^{N_{\mathcal{C}_1}} \sim \mathcal{P}_{\mathcal{C}_1} = \mathbf{N}\left(\mu_{\mathcal{C}_1}, \sigma_{\mathcal{C}_1}\right) $ and $\left\{X_i\right\}_{i=0}^{N_{\mathcal{C}_2}} \sim \mathcal{P}_{\mathcal{C}_2} = \mathbf{N}\left(\mu_{\mathcal{C}_2}, \sigma_{\mathcal{C}_2}\right)$. This assumption holds on the condition that the data variables are composed of a wealth of statistically independent random factors on graph benchmarks.
\end{assumption}
Then, as the thorough distributions of candidate categories $\mathcal{C}_1$ and ${\mathcal{C}_2}$ are agnostic, we demonstrate the probability of error for task $\mathcal{T}^b$ as follows:
\begin{lemma}
\label{lem:1}
Given the unknown mean vectors $\mu_{\mathcal{C}_1}$ and $\mu_{\mathcal{C}_2}$ and $N_S = N_{\mathcal{C}_1} + N_{\mathcal{C}_2}$ labeled training samples, we can recap the optimal Bayes decision rule to construct the decision boundary based on a plain 0-1 loss function and derive the probability of error as
\begin{equation} \label{eq:r0-1}
    \mathcal{R}_{\mathcal{T}^b} \left(N_S, D\right) = \int_{\kappa\left(D\right)}^\infty \frac{1}{\sqrt{2\pi}} \cdot e^{-\frac{\rho^2}{2}}d\rho,
\end{equation}
where $D$ denotes the dimensionality of graph representations learned by a neural network and 
\begin{equation}
    \kappa\left(D\right) = \frac{\sum_{\nu=1}^D\frac{1}{\nu}}{\sqrt{\frac{D}{N_S} + \left(1+ \frac{1}{N_S}\right) \cdot \sum_{\nu=1}^D\frac{1}{\nu}}}.
\end{equation}
\end{lemma}
Lemma \ref{lem:1} states that the probability of error is directly related to the training sample size $N_S$ and the representation dimensionality $D$. According to the experimental setting of graph benchmarks, $N_S$ is fixed among comparisons. The proposed HTML is a plug-and-play method that fits various GCL methods without mandating a shift in the dimensionality of the learned representations, thereby preserving a constant $D$. Therefore, we attest that the theoretical analyses on the Bayes errors of the canonical GCL method and HTML can be conducted in a fair manner. Considering the complexity gap between the cross-entropy loss and the 0-1 loss, we introduce an approach for estimating the bounds of Bayes error based on divergence \cite{DBLP:journals/pami/JainDM00} by following the principle of Mahalanobis distance measure \cite{de2000mahalanobis} as follows:
\begin{lemma}
\label{lem:2}
Given the mean vectors $\mu_{\mathcal{C}_1}$, $\mu_{\mathcal{C}_2}$ and the covariance matrices $\sigma_{\mathcal{C}_1}$, $\sigma_{\mathcal{C}_2}$ for the hypothesis Gaussian distributions $\mathcal{P}_{\mathcal{C}_1}$ and $\mathcal{P}_{\mathcal{C}_2}$, respectively, we denote $\bar{\sigma}$ be the non-singular average covariance matrix over $\sigma_{\mathcal{C}_1}$ and $\sigma_{\mathcal{C}_2}$, i.e., $\bar{\sigma}=P({\mathcal{C}_1})\cdot\sigma_{\mathcal{C}_1} + P({\mathcal{C}_2})\cdot\sigma_{\mathcal{C}_2}$, where $P({\mathcal{C}_1})$ and $P({\mathcal{C}_2})$ are the probability densities corresponding to the prior distributions $\mathcal{P}_{\mathcal{C}_1}$ and $\mathcal{P}_{\mathcal{C}_2}$. We can conduct the divergence estimation on the upper bound of Bayes error $\mathcal{R}_{\mathcal{T}^b}$ by
\begin{equation} \label{eq:beub}
    \mathcal{R}_{\mathcal{T}^b} \leq \frac{2P({\mathcal{C}_1})P({\mathcal{C}_2})}{1+P({\mathcal{C}_1})P({\mathcal{C}_2})\cdot\left[(\mu_{\mathcal{C}_1}-\mu_{\mathcal{C}_2})^\top\bar{\sigma}^{-1}(\mu_{\mathcal{C}_1}-\mu_{\mathcal{C}_2})\right]},
\end{equation}
\end{lemma}
\begin{table*}[ht]
\small
\setlength{\tabcolsep}{5.5pt}
\caption{Comparing classification accuracy (\%) with unsupervised representation learning setting. We report the average accuracy in the last column, and the top three results are \textbf{bolded}.}
\label{unsuper}
\centering
\begin{tabular}{cccccccccc}
\hline
 Methods & NCI1 & PROTEINS & DD & MUTAG & COLLAB & RDT-B & RDT-M5K & IMDB-B & AVG. \\
\hline
 GL & $-$ & $-$ & $-$ & 81.66\tiny{$\pm$2.11} & $-$ & 77.34\tiny{$\pm$0.18} & 41.01\tiny{$\pm$0.17 }&65.87\tiny{$\pm$0.98} & 66.47 \\
WL & \textbf{80.01}\tiny{$\pm$0.50} & 72.92\tiny{$\pm$0.56} & $-$ & 80.72\tiny{$\pm$3.00} & $-$ & 68.82\tiny{$\pm$0.41} & 46.06\tiny{$\pm$0.21} & \textbf{72.30}\tiny{$\pm$3.44} & 70.14 \\
DGK & \textbf{80.31}\tiny{$\pm$0.46} & 73.30\tiny{$\pm$0.82} & $-$ & 87.44\tiny{$\pm$2.72} & $-$ & 78.04\tiny{$\pm$0.39} & 41.27\tiny{$\pm$0.18} & 66.96\tiny{$\pm$0.56} & 71.22 \\
\hline

InfoGraph&76.20\tiny{$\pm$1.06} & 74.44\tiny{$\pm$0.31}&72.85\tiny{$\pm$1.78} & \textbf{89.01}\tiny{$\pm$1.13} & 70.65\tiny{$\pm$1.13}&82.50\tiny{$\pm$1.42}&53.46\tiny{$\pm$1.03} & \textbf{73.03}\tiny{$\pm$0.87}&74.02 \\
GraphCL&77.87\tiny{$\pm$0.41}&74.39\tiny{$\pm$0.45} & \textbf{78.62}\tiny{$\pm$0.40}&86.80\tiny{$\pm$1.34} & \textbf{71.36}\tiny{$\pm$1.15} & 89.53\tiny{$\pm$0.84} & \textbf{55.99}\tiny{$\pm$0.28} & 71.14\tiny{$\pm$0.44}& \textbf{75.71} \\
JOAO&78.07\tiny{$\pm$0.47} & \textbf{74.55}\tiny{$\pm$0.41}&77.32\tiny{$\pm$0.54}&87.35\tiny{$\pm$1.02}&69.50\tiny{$\pm$0.36}&85.29\tiny{$\pm$1.35}&55.74\tiny{$\pm$0.63}&70.21\tiny{$\pm$3.08}&74.75 \\
JOAOv2&78.36\tiny{$\pm$0.53}&74.07\tiny{$\pm$1.10} & 77.40\tiny{$\pm$1.15} & 87.67\tiny{$\pm$0.79}&69.33\tiny{$\pm$0.34} & 86.42\tiny{$\pm$1.45} & \textbf{56.03}\tiny{$\pm$0.27}&70.83\tiny{$\pm$0.25} & 75.01 \\
AD-GCL& 73.91\tiny{$\pm$0.77}& 73.28\tiny{$\pm$0.46}& 75.79\tiny{$\pm$0.87}&\textbf{88.74}\tiny{$\pm$1.85}& \textbf{72.02}\tiny{$\pm$0.56}& \textbf{90.07}\tiny{$\pm$0.85}&54.33\tiny{$\pm$0.32} &70.21\tiny{$\pm$0.68} & 74.79\\
RGCL& 78.14\tiny{$\pm$1.08}& \textbf{75.03}\tiny{$\pm$0.43}& \textbf{78.86}\tiny{$\pm$0.48} & 87.66\tiny{$\pm$1.01}& 70.92\tiny{$\pm$0.65}& \textbf{90.34}\tiny{$\pm$0.58}& \textbf{56.38}\tiny{$\pm$0.40}& \textbf{71.85}\tiny{$\pm$0.84}& \textbf{76.15}\\

\hline
GraphCL+HTML & \textbf{78.72}\tiny{$\pm$0.72} & \textbf{74.95}\tiny{$\pm$0.33 }& \textbf{78.60}\tiny{$\pm$0.18}& \textbf{88.91}\tiny{$\pm$1.85}& \textbf{71.60}\tiny{$\pm$0.81}& \textbf{90.66}\tiny{$\pm$0.64}& 55.95\tiny{$\pm$0.36}& 71.68\tiny{$\pm$0.42}& \textbf{76.38} \\
\textbf{$\Delta$} &+0.85&+0.56 &-0.02&+2.11&+0.24&+1.13&-0.04&+0.54&+0.67 \\
\hline
\end{tabular}
\end{table*}

Holding Lemma \ref{lem:2}, we can deduce the upper bound of the error gap between the canonical GCL method and HTML as follows:
\begin{proposition}
\label{pro:1}
 \cite{tumer1996linear} states that the Bayes optimum decision is the loci of all samples $X^{\star}$ satisfying $P({\mathcal{C}_1}|X^{\star})=P({\mathcal{C}_2}|X^{\star})$. Then, from the perspective of classification, we can derive that the predicted decision boundary of the canonical GCL model $f$ is the loci of the available samples $X^A$ satisfying $f(\mathcal{C}_1|X^A)=f(\mathcal{C}_2|X^A)$, and there exists a constant gap $X^{\Delta}$, i.e., $\exists X^{\Delta} = X^A - X^\star$.
\end{proposition}
Considering Proposition \ref{pro:1}, we deduce that
\begin{corollary}
\label{cor:1}
Given an independent and sole GCL model $f$ for classification and the available training samples $X^A$, we can get $f(\mathcal{C}|X^A) = P(\mathcal{C}|X^A) + \mathcal{E}(\mathcal{C}, f, X^A)$, where $\mathcal{E}(\cdot, \cdot, \cdot)$ is the inherent error. Following \cite{tumer1996linear}, $\mathcal{E}(\mathcal{C}, f, X^A)$ is implemented by
\begin{equation} \label{eq:inerror}
    \mathcal{E}(\mathcal{C}, f, X^A) = \Phi_{\mathcal{C}} + \Psi_{\mathcal{C}, f, X^A},
\end{equation}
where $\Phi_{\mathcal{C}}$ is constant for the category $\mathcal{C}$, while $\Psi_{\mathcal{C}, f, X^A}$ is additionally associated with the model $f$ and the training samples $X^A$.
\end{corollary}
Considering Equation \ref{eq:inerror}, we can arbitrarily discard $\Phi_{\mathcal{C}}$ during the deduction of the error gap between the canonical GCL model and HTML since the available training samples and categories are constant on graph benchmarks, such that $\Phi_{\mathcal{C}}$ is constant, and only $\Psi_{\mathcal{C}, f, X^A}$ varies along with the change of model.

The proposed HTML can be treated as an implicit multi-loss ensemble learning model \cite{DBLP:journals/corr/abs-2109-14433}, and the features learned by back-propagating detached losses are contained by multiple-dimensional subsets of the ultimate representations. The classification weights, following the linear probing scheme, perform the implicitly weighted ensemble for representations. Therefore, HTML can be decomposed into the canonical GCL model $f$ and the plug-and-play hierarchical topology isomorphism expertise learning model $f_{\star}$, such that we derive the error gap between the candidate models as follows:
\begin{theorem}
\label{thm:beghtml}
Given the decomposed HTML models $f$ and $f_{\star}$ for classification, the Bayes error gap between the proposed HTML and the canonical GCL model can be approximated by
\begin{equation}
    \begin{aligned}
        &\mathcal{R}_{\mathcal{T}^b}(f) - \mathcal{R}_{\mathcal{T}^b}(\left\{f, f_{\star}\right\}) = \\
    &\frac{3\delta_{\Psi_{\mathcal{C}_1, f,X^A}}^2 + 3 \delta_{\Psi_{\mathcal{C}_2, f,X^A}}^2 - \delta_{\Psi_{\mathcal{C}_1, f_{\star}, X^A}}^2 - \delta_{\Psi_{\mathcal{C}_2, f_{\star}, X^A}}^2}{8|\nabla_{X^A=X^{\star}} P({\mathcal{C}_1}|X^A) - \nabla_{X^A=X^{\star}} P({\mathcal{C}_2}|X^A)|}.
    \end{aligned}
\end{equation}
\end{theorem}
Refer to \textbf{Appendix \ref{app:beghtmlproof}} for the corresponding proofs.According to the experimental setting of graph benchmarks, the inherent error $\Psi$ is associated and only associated with models $f$ and $f_{\star}$, and both $\mathcal{R}_{\mathcal{T}^b}(f)$ and $\mathcal{R}_{\mathcal{T}^b}(\left\{f, f_{\star}\right\})$ are associated with $\Psi$, such that the Bayes error gap $\mathcal{R}_{\mathcal{T}^b}(f) - \mathcal{R}_{\mathcal{T}^b}(\left\{f, f_{\star}\right\})$ is solely dependent to the candidate models $f$ and $f_{\star}$, which further substantiates that the theoretical analyses are imposed in an impartial scenario. Holding the empirically solid hypothesis that the differences of inherent errors, i.e., $\left<\Psi_{\mathcal{C}_1, f,X^A}, \Psi_{\mathcal{C}_1, f_{\star},X^A}\right>$ and $\left<\Psi_{\mathcal{C}_2, f,X^A}, \Psi_{\mathcal{C}_2, f_{\star},X^A}\right>$, are bounded, that is, the performance gap of the GCL model and HTML is not dramatically large, we can derive the following: 
\begin{corollary}
\label{cor:2}
Given the candidate models $f$ and $f_{\star}$, the available training samples $X^A$, and the labels $\mathcal{C}$ of target downstream task $\mathcal{T}^b$, we suppose that the corresponding inherent errors are bounded, i.e., $-\epsilon_f \leq \Psi_{\mathcal{C}, f, X^A} \leq \epsilon_f$ and $-\epsilon_{f_{\star}} \leq \Psi_{\mathcal{C}, f_{\star}, X^A} \leq \epsilon_{f_{\star}}$. If $\epsilon_{f_{\star}} \leq \sqrt{3}\cdot\epsilon_f$ holds, the positive definiteness of the Bayes error gap between $\mathcal{R}_{\mathcal{T}^b}(f)$ and $\mathcal{R}_{\mathcal{T}^b}(\left\{f, f_{\star}\right\})$ can be satisfied, i.e., we can establish that $\mathcal{R}_{\mathcal{T}^b}(f) - \mathcal{R}_{\mathcal{T}^b}(\left\{f, f_{\star}\right\}) \geq 0$.
\end{corollary}

Refer to \textbf{Appendix \ref{app:cor2proof}} for the proof.The derivations in Theorem \ref{thm:beghtml} and Corollary \ref{cor:2} generally fit the multi-dimensional latent space~ \cite{tumer1996linear}, such that we can state that compared with the canonical GCL methods, the proposed HTML acquires the relatively tighter Bayes error upper bound, which substantiates that the performance rises of GCL methods incurred by HTML are theoretical solid.
\begin{table}[ht]
\scriptsize
\setlength{\tabcolsep}{4.pt}
\caption{Comparing classification accuracy (\%) with unsupervised representation learning setting. We report the average accuracy in the last column and the improvement compared to the baseline, namely RGCL, in the last row. \textbf{Bold} indicates the best result, and \underline{underline} indicates the second best result.}
\label{unsuper_rgcl}
\centering
\begin{tabular}{cccccc}
 \hline
 Methods & NCI1& MUTAG & COLLAB  & IMDB-B & AVG. \\
\hline
 
No Pre-Train&65.40\tiny{$\pm$0.17} & 87.39\tiny{$\pm$1.09} & 65.29\tiny{$\pm$0.16}& 69.37\tiny{$\pm$0.37}&71.86 \\
InfoGraph&76.20\tiny{$\pm$1.06} & \textbf{89.01}\tiny{$\pm$1.13} & 70.65\tiny{$\pm$1.13} & \textbf{73.03}\tiny{$\pm$0.87}&\underline{77.22} \\
GraphCL&77.87\tiny{$\pm$0.41}&86.80\tiny{$\pm$1.34} & \underline{71.36\tiny{$\pm$1.15}} & 71.14\tiny{$\pm$0.44}& 76.79 \\
AD-GCL& 73.91\tiny{$\pm$0.77}&\underline{88.74\tiny{$\pm$1.85}}& \textbf{72.02}\tiny{$\pm$0.56}&70.21\tiny{$\pm$0.68} & 76.22\\
RGCL& \underline{78.14\tiny{$\pm$1.08}}& 87.66\tiny{$\pm$1.01}& 70.92\tiny{$\pm$0.65}& 71.85\tiny{$\pm$0.84}& 77.14\\

\hline 
RGCL+HTML & \textbf{78.38}\tiny{$\pm$0.67} & 88.72\tiny{$\pm$1.90}& 71.19\tiny{$\pm$1.13}& \underline{72.17\tiny{$\pm$0.86}}& \textbf{77.62} \\
$\Delta$ &+0.24&+1.06&+0.27&+0.32&+0.48 \\
\hline
\end{tabular}
\vspace{-0.15cm}
\end{table}
\begin{table*}
\small
\caption{Comparing ROC-AUC scores (\%) on downstream graph classification tasks with transfer learning setting.}
\label{tab:transfer}
\setlength{\tabcolsep}{5.5pt}
\centering
\begin{tabular}{ccccccccccc}
\hline
Methods &  BBBP &Tox21&  ToxCast & SIDER & ClinTox &  MUV &  HIV &  BACE &  GAIN. \\
\hline
No Pre-Train& 65.8\tiny{$\pm$4.5}& 74.0\tiny{$\pm$0.8} & 63.4\tiny{$\pm$0.6} & 57.3\tiny{$\pm$1.6} & 58.0\tiny{$\pm$4.4} &71.8\tiny{$\pm$2.5} & 75.3\tiny{$\pm$1.9} & 70.1\tiny{$\pm$5.4} & -\\
\hline
Infomax& 68.8\tiny{$\pm$0.8}& 75.3\tiny{$\pm$0.5} & 62.7\tiny{$\pm$0.4} & 58.4\tiny{$\pm$0.8} &69.9\tiny{$\pm$3.0} &75.3\tiny{$\pm$2.5} & 76.0\tiny{$\pm$0.7} & 75.9\tiny{$\pm$1.6} & 3.3\\
EdgePred& 67.3\tiny{$\pm$2.4}& \textbf{76.0}\tiny{$\pm$0.6} & \textbf{64.1}\tiny{$\pm$0.6} &60.4\tiny{$\pm$0.7} &64.1\tiny{$\pm$3.7} &74.1\tiny{$\pm$2.1} & 76.3\tiny{$\pm$1.0} & \textbf{79.9}\tiny{$\pm$0.9} & 3.3\\
AttrMasking& 64.3\tiny{$\pm$ 2.8}& \textbf{76.7}\tiny{$\pm$ 0.4}& \textbf{64.2}\tiny{$\pm$ 0.5}& \textbf{61.0}\tiny{$\pm$0.7}&
71.8\tiny{$\pm$4.1}&
74.7\tiny{$\pm$1.4}&
77.2\tiny{$\pm$1.1}&
\textbf{79.3}\tiny{$\pm$1.6}&4.1 \\
ContextPred& 68.0\tiny{$\pm$2.0}& \textbf{75.7}\tiny{$\pm$0.7}& \textbf{63.9}\tiny{$\pm$0.6} &60.9\tiny{$\pm$0.6} & 65.9\tiny{$\pm$3.8}&\textbf{75.8}\tiny{$\pm$1.7}&77.3\tiny{$\pm$1.0} &\textbf{79.6}\tiny{$\pm$1.2} &3.9\\
GraphCL& 69.68\tiny{$\pm$0.67}& 73.87\tiny{$\pm$0.66}&
62.40\tiny{$\pm$0.57}&
60.53\tiny{$\pm$0.88}&
75.99\tiny{$\pm$2.65}& 69.80\tiny{$\pm$2.66}&
\textbf{78.47}\tiny{$\pm$1.22}&
75.38\tiny{$\pm$1.44} & 3.77\\

JOAO & \textbf{70.22}\tiny{$\pm$0.98} &	74.98\tiny{$\pm$0.29} &
62.94\tiny{$\pm$0.48}&
59.97\tiny{$\pm$0.79}&	
\textbf{81.32}\tiny{$\pm$2.49}&
71.66\tiny{$\pm$1.43}&
76.73\tiny{$\pm$1.23}&
77.34\tiny{$\pm$0.48}&
\textbf{4.90}\\
AD-GCL&	68.26\tiny{$\pm$1.03}&	
73.56\tiny{$\pm$0.72}&	
63.10\tiny{$\pm$0.66}&	
59.24\tiny{$\pm$0.86}&	
77.63\tiny{$\pm$4.21}&	
74.94\tiny{$\pm$2.54}&	
75.45\tiny{$\pm$1.28}&
75.02\tiny{$\pm$1.88}&
3.90\\
RGCL&\textbf{71.42}\tiny{$\pm$0.66}&	
75.20\tiny{$\pm$0.34}&	
63.33\tiny{$\pm$0.17}&	
\textbf{61.38}\tiny{$\pm$0.61}&	
\textbf{83.38}\tiny{$\pm$0.91}&	
\textbf{76.66}\tiny{$\pm$0.99}&	
\textbf{77.90}\tiny{$\pm$0.80}&
76.03\tiny{$\pm$0.77}&
\textbf{6.16}\\
\hline
GraphCL+HTML& \textbf{70.94}\tiny{$\pm$0.65}&
74.69\tiny{$\pm$0.28}&
63.73\tiny{$\pm$0.30}&
\textbf{61.32}\tiny{$\pm$0.59}&
\textbf{84.03}\tiny{$\pm$0.12}&
\textbf{78.05}\tiny{$\pm$0.94}&
\textbf{78.68}\tiny{$\pm$0.61}&
77.24\tiny{$\pm$0.90}& \textbf{6.59}\\
\textbf{$\Delta$} &+1.26&+0.82 &+1.33&+0.79&+8.04&+8.25&+0.21&+1.86& +2.82\\
\hline
\end{tabular}
\end{table*}

\section{Experiments}
In this section, we experimentally demonstrate the validity of HTML on 17 real-world datasets and analyze the effects of each module of HTML. We introduce the datasets and experimental settings in \textbf{Appendix \ref{app:data}}.

\subsection{Performance on Real-World Graph Representation Learning Tasks}

\textbf{Benchmarking on unsupervised learning towards regular graphs}. Table \ref{unsuper} compares the classification accuracy between our proposed HTML and the SOTA methods under the unsupervised representation learning setting. As can be seen, HTML achieves the top three on almost all datasets and ranks first in average accuracy. This verifies our conclusion in \textbf{Section} \ref{sec:intro}, i.e., guiding GNNs to learn topology-level information can improve the capacity of GCL methods to learn discriminative representations. Since HTML is implemented based on GraphCL, we compare the performance between HTML and GraphCL and show the increase in the last row in Table \ref{unsuper}. According to Table \ref{unsuper}, GraphCL+HTML shows improvements over GraphCL with an average increase of 0.67\%. We perform the significance test, i.e., paired t-test \cite{ross2017paired}, and observe that GraphCL+HTML consistently shows a difference in performance compared to GraphCL with the P-value below 0.05, i.e., 0.03.

\begin{figure}
        \centering
		\begin{minipage}[h]{0.22\textwidth}
			
                \centering
			\includegraphics[width=1\textwidth]{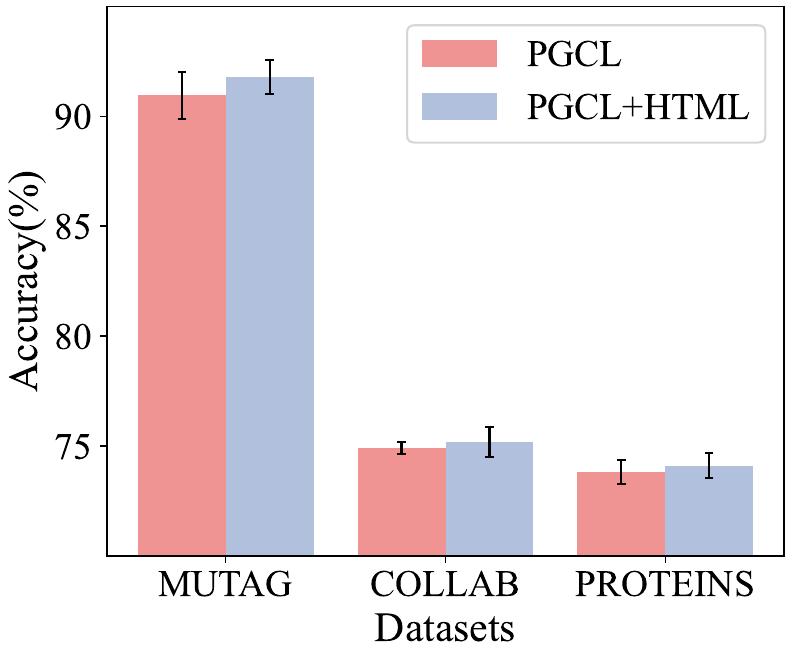}
			\caption{Empirical validation of the improvement robustness of HTML towards the variance of baselines.}
			\label{fig:pgclhtml}
		\end{minipage}
  \
             \begin{minipage}[h]{0.22\textwidth}
                \centering
			\includegraphics[width=1\textwidth]{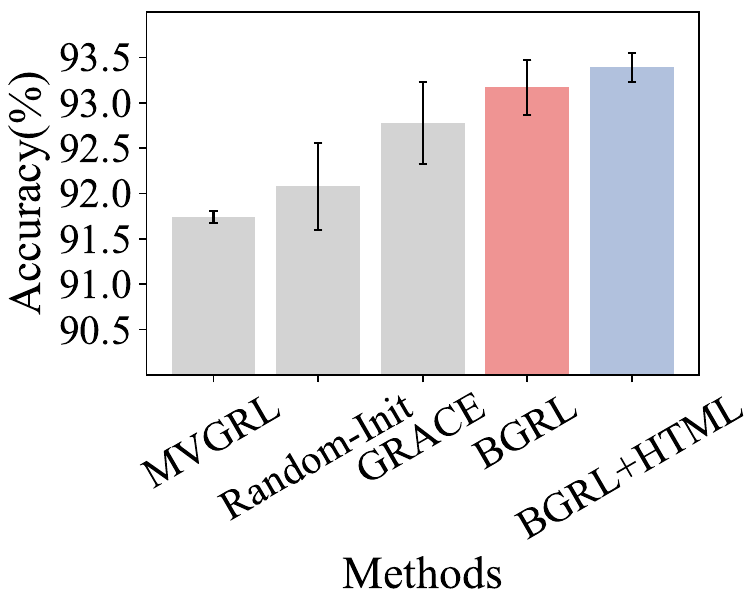}
			\caption{Comparisons on the large-scale Amazon Photos dataset for node classification task.}
			\label{fig:large}
		\end{minipage}
	\vspace{-0.25cm}
\end{figure}

\textbf{Benchmarking on unsupervised learning with other baselines}. HTML is a plug-and-play method that fits and improves various baselines well. To validate the generality of the performance improvement derived by introducing HTML, we conduct further comparisons on other backbones. We implement HTML with RGCL as the backbone, and the results on four representative datasets are shown in Table \ref{unsuper_rgcl}. The proposed HTML empowers RGCL to achieve the highest average accuracy. Besides, as shown in Figure \ref{fig:pgclhtml}, PGCL+HTML can consistently outperform PGCL over different benchmarks, demonstrating that the improvement brought about by HTML is robust to the variance of the adopted baselines.

\textbf{Benchmarking on unsupervised learning towards large-scale graphs}. To validate the effectiveness of our proposed HTML on large-scale graphs, we implement HTML as a plug-and-play component on BGRL and conduct experiments on the Amazon Photo dataset with 7,650 nodes and 119,081 edges. BGRL is a SOTA model for node classification tasks for unsupervised learning towards large-scale graphs, and the comparisons between BGRL+HTML and benchmarks are shown in Figure \ref{fig:large}. Experimental results show that BGRL+HTML achieves the best performance among benchmarks, which validates the generalization of HTML.

\textbf{Benchmarking on transfer learning}. Table \ref{tab:transfer} compares our proposed HTML with the SOTA methods under the transfer learning setting. The last column of this table shows the average improvement compared to the method without pre-training. HTML achieves the highest boost among the compared methods on average, supporting the transferability of our approach. Specifically, we engage in a comparative analysis between HTML and the principal baseline, i.e., GraphCL, in the last row of Table \ref{tab:transfer}. It is worth noting that HTML achieves significant performance boosts in all eight datasets, with an average increase of 2.82\%. Compared to the SOTA approach, i.e., RGCL, HTML derives an average performance improvement of 0.43\%. Overall, the results demonstrate the valid transferability of HTML. We perform the significance paired t-test \cite{ross2017paired} and observe that GraphCL+HTML shows a difference in performance compared to GraphCL with the P-value below 0.05, i.e., 0.047.

\begin{figure}
  \centering
  \includegraphics[scale=0.43]{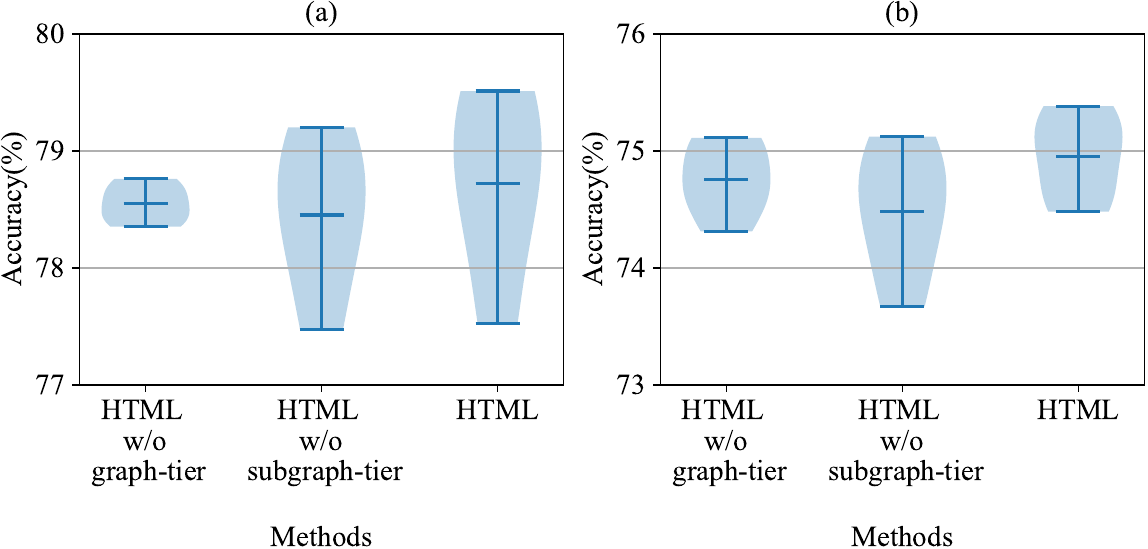}
  \caption{Ablation study on (a) NCI1 and (b) PROTEINS.}
\label{fig:ablation}

\end{figure}

\subsection{In-Depth Analysis and Discussion}
\begin{figure}
  \centering
  \includegraphics[scale=0.35]{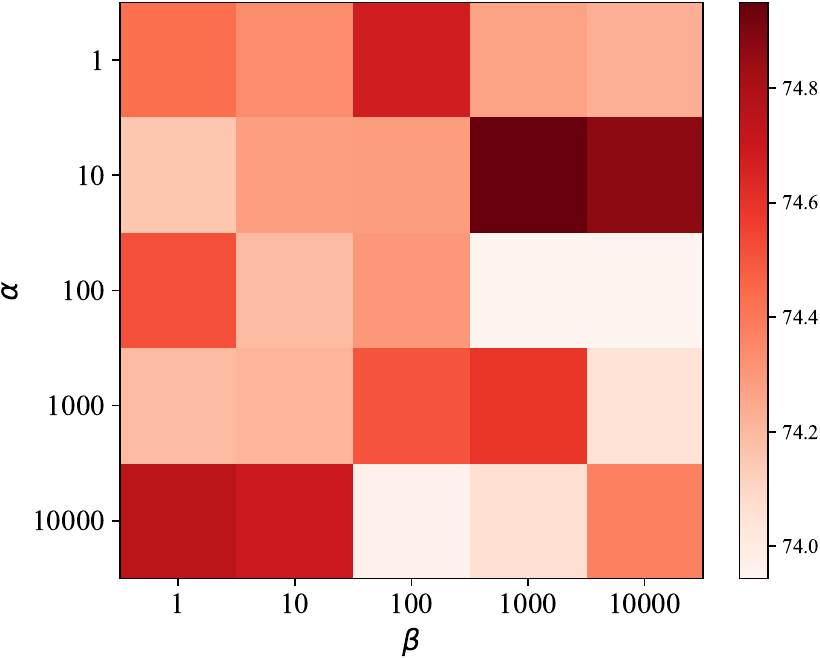}
  \caption{Accuracy (\%) with different hyper-parameters.}
\label{fig:hpyer}
\end{figure}

\textbf{Ablation study}.
To verify the effectiveness of each module in HTML, we conduct ablation studies on NCI1 and PROTEINS datasets, shown in Figure \ref{fig:ablation}. We conduct experiments in three settings: HTML, HTML without graph-tier expertise, and HTML without subgraph-tier expertise. We perform five experiments for each setting with five different seeds, and the results are presented as a violin plot. The width of the violin represents the distribution of results, and the blue horizontal line in the middle indicates the mean value of the five results. It can be observed that removing any of the modules from the HTML causes a certain drop in performance. However, the ablation models can still outperform the principle baseline, i.e., GraphCL, which confirms our previous conjecture that GNN-based GCL models lack topology-level information and suggests that our proposed graph-tier expertise and subgraph-tier expertise are both contributing to the GCL model. Figure \ref{fig:ablation} reveals that adding subgraph-tier expertise leads to a more significant boost to the GCL model and usually has a more stable result under different seeds, indicating that fine-grained expertise is essential for improving model performance and stability.\\
\textbf{Hyper-parameter experiment}.
Appropriate adjustment of the introduction of hierarchical topology isomorphism expertise can better improve the model performance, so we conduct hyperparametric experiments on the PROTEINS dataset for $\alpha$ and $\beta$, and the results are shown in Figure \ref{fig:hpyer}. $\alpha$ and $\beta$ are range from $\{1,1\times10^1,1\times10^2,1\times10^3,1\times10^4\}$. $\alpha$ controls the weight of the graph-tier topology isomorphism expertise module, and $\beta$ controls the weight of the subgraph-tier topology isomorphism expertise module. As shown in Figure \ref{fig:hpyer}, the model achieves the highest accuracy at $\alpha=10$ and $\beta=1000$. It is easy to find that when $\beta$ takes low values, the higher $\alpha$ achieves better results, while when $\beta$ takes high values, the lower $\alpha$ achieves better results. We argue that this phenomenon may be because the topology isomorphism expertise at graph-tier and subgraph-tier may have certain complementarity.

\section{Conclusions}
By observing the motivating experiments, we derive the empirical conclusion that GNN-based GCL methods can barely learn the hierarchical topology isomorphism expertise during training. Thus, the corresponding expertise is complementary to GCL methods. To this end, we propose HTML, which introduces the hierarchical topology isomorphism expertise into the GCL paradigm. Theoretically, we provide validated analyses to demonstrate that compared with conventional GCL methods, HTML derives the tighter Bayes error upper bound. Empirically, benchmark comparisons prove that HTML yields significant performance boosts.


\section*{Acknowledgements}
The authors would like to thank the editors and reviewers for their valuable comments. This work is supported by the 2022 Special Research Assistant Grant project, No. E3YD5901, the CAS Project for Young Scientists in Basic Research, Grant No. YSBR-040, the Youth Innovation Promotion Association CAS, No. 2021106, and the China Postdoctoral Science Foundation, No. 2023M743639.

\newpage
\bibliography{aaai24}

\newpage
\appendix

\onecolumn
\section{Theoretical Proofs} \label{app:proof}
To confirm the correctness and integrity of the Theorem and Corollary, we elaborate on the following proof.

\subsection{Proof for Theorem 5.6} \label{app:beghtmlproof}
Primarily, considering Proposition 5.4, we deduce that the upper bound of Bayes error can be further defined by
\begin{equation}
\mathcal{R}_{\mathcal{T}^b}(f) = \mathcal{R}_{\mathcal{T}^b} + \mathcal{R}_{CLS},
\label{eq:rextra}
\end{equation}
where $\mathcal{R}_{\mathcal{T}^b}$ denotes the inherent Bayes error brought by the candidate graph representation learning model, and $\mathcal{R}_{CLS}$ denotes the error brought by the classification head.

Accordingly, following the deduction in Proposition 5.4, we acquire $f(\mathcal{C}_{1}|X^A) = f(\mathcal{C}_{2}|X^A)$, which can be further transformed into the below form based on the implementation in Corollary 5.5 as follows:
\begin{equation}
\begin{aligned}
    f(\mathcal{C}_{1}|X^A) &= f(\mathcal{C}_{2}|X^A)\\
    P(\mathcal{C}_{1}|X^A) + \mathcal{E}(\mathcal{C}_{1}, f, X^A) &= P(\mathcal{C}_{2}|X^A) + \mathcal{E}(\mathcal{C}_{2}, f, X^A)\\
    P\left(\mathcal{C}_{1}|X^\star + X^{\Delta}\right) + \mathcal{E}(\mathcal{C}_{1}, f, X^A) &= P\left(\mathcal{C}_{2}|X^\star + X^{\Delta}\right) + \mathcal{E}(\mathcal{C}_{2}, f, X^A).
\end{aligned}
\label{eq:proofeq1}
\end{equation}
Thus, using a linear function to approximate $P\left(\mathcal{C}|X^A\right)$ around $X^\star$ is achievable, and we acquire
\begin{equation}
\left\{
    \begin{aligned}
         P\left(\mathcal{C}_{1}|X^\star + X^{\Delta}\right)	&\thickapprox P\left(\mathcal{C}_{1}|X^\star\right) + X^{\Delta}\nabla P\left(\mathcal{C}_{1}|X^\star\right)\\
    P\left(\mathcal{C}_{2}|X^\star + X^{\Delta}\right)	&\thickapprox P\left(\mathcal{C}_{2}|X^\star\right) + X^{\Delta}\nabla P\left(\mathcal{C}_{2}|X^\star\right).
    \end{aligned}
    \label{eq:proofeq2}
\right.
\end{equation}
Considering the approximate equations defined in Equation \ref{eq:proofeq2} and the equity constraint defined in Proposition 5.4, Equation \ref{eq:proofeq1} can be transformed into
\begin{equation}
    \begin{aligned}
        P\left(\mathcal{C}_{1}|X^\star + X^{\Delta}\right) + \mathcal{E}(\mathcal{C}_{1}, f, X^A) &= P\left(\mathcal{C}_{2}|X^\star + X^{\Delta}\right) + \mathcal{E}(\mathcal{C}_{2}, f, X^A)\\
        P\left(\mathcal{C}_{1}|X^\star\right) + X^{\Delta}\nabla P\left(\mathcal{C}_{1}|X^\star\right) + \mathcal{E}(\mathcal{C}_{1}, f, X^A) &= P\left(\mathcal{C}_{2}|X^\star\right) + X^{\Delta}\nabla P\left(\mathcal{C}_{2}|X^\star\right) + \mathcal{E}(\mathcal{C}_{2}, f, X^A)\\
        X^{\Delta}\nabla P\left(\mathcal{C}_{1}|X^\star\right) + \mathcal{E}(\mathcal{C}_{1}, f, X^A) &= X^{\Delta}\nabla P\left(\mathcal{C}_{2}|X^\star\right) + \mathcal{E}(\mathcal{C}_{2}, f, X^A).
    \end{aligned}
    \label{eq:proofeq3}
\end{equation}
Accordingly, the analytical solution of Equation \ref{eq:proofeq3} can be derive as follows:
\begin{equation}
    \begin{aligned}
        X^{\Delta}\nabla P\left(\mathcal{C}_{1}|X^\star\right) + \mathcal{E}(\mathcal{C}_{1}, f, X^A) &= X^{\Delta}\nabla P\left(\mathcal{C}_{2}|X^\star\right) + \mathcal{E}(\mathcal{C}_{2}, f, X^A)\\
        \left[\nabla P\left(\mathcal{C}_{1}|X^\star\right) - \nabla P\left(\mathcal{C}_{2}|X^\star\right)\right]X^{\Delta} &= \mathcal{E}(\mathcal{C}_{2}, f, X^A) - \mathcal{E}(\mathcal{C}_{1}, f, X^A)\\
        X^{\Delta} &= \frac{\mathcal{E}(\mathcal{C}_{1}, f, X^A)-\mathcal{E}(\mathcal{C}_{2}, f, X^A)}{\nabla P\left(\mathcal{C}_{2}|X^\star\right)-\nabla P\left(\mathcal{C}_{1}|X^\star\right)}.
    \end{aligned}
    \label{eq:proofeq4}
\end{equation}
Holding Proposition 5.4, we can intuitively deduce the formalized implementation of $\mathcal{R}_{CLS}$ as follows:
\begin{equation}
    \mathcal{R}_{CLS} = \int_{-\infty}^{+\infty}\int_{X^\star}^{X^\star+X^{\Delta}}|P\left(\mathcal{C}_{1}|X^A\right) - P\left(\mathcal{C}_{2}|X^A\right)|P(X^{\Delta})dX^AdX^{\Delta},
    \label{eq:proofeq5}
\end{equation}
where $P(X^{\Delta})$ presents the probability density for $X^{\Delta}$.

Equation \ref{eq:proofeq1} and Equation \ref{eq:proofeq2} jointly provide the approximation analyses on $P\left(\mathcal{C}|X^A\right)$ and derive the corresponding approximation implementations, such that we can straightforwardly deduce that
\begin{equation}
    \begin{aligned}
        \mathcal{R}_{CLS} &= \int_{-\infty}^{+\infty}\int_{X^\star}^{\left(X^\star+X^{\Delta}\right)}(X^A-X^\star)|\nabla P\left(\mathcal{C}_{2}|X^\star\right)-\nabla P\left(\mathcal{C}_{1}|X^\star\right)|P(X^{\Delta})dX^AdX^{\Delta}\\
    &=\int_{-\infty}^{+\infty}\frac{1}{2}\left({X^{\Delta}}\right)^2|\nabla P\left(\mathcal{C}_{2}|X^\star\right)-\nabla P\left(\mathcal{C}_{1}|X^\star\right)|P(X^{\Delta})dX^{\Delta}\\
    &=\frac{1}{2}|\nabla P\left(\mathcal{C}_{2}|X^\star\right)-\nabla P\left(\mathcal{C}_{1}|X^\star\right)|\delta_{\Psi_{\mathcal{C}, f,X^\Delta}}^2.
    \end{aligned}
    \label{eq:proofeq6}
\end{equation}
Following the derivation in Corollary~5.5, we have $\mathcal{E}(\mathcal{C}, f, X^A) = \Phi_{\mathcal{C}}+\Psi_{\mathcal{C}, f, X^A}(X^A)$. Holding a prior assumption that the bias of $X^\Delta$ is consistently trivial, i.e., the bias of $X^\Delta$ is 0 for the available training samples $X^A$, such that following Equation \ref{eq:proofeq4}, the variance of $X^\Delta$ can be derive by 
\begin{equation}
    \delta_{\Psi_{\mathcal{C}, f,X^\Delta}}^2=\frac{\delta_{\Psi_{\mathcal{C}_1, f,X^A}}^2 + \delta_{\Psi_{\mathcal{C}_2, f,X^A}}^2}{|\nabla P\left(\mathcal{C}_{2}|X^\star\right)-\nabla P\left(\mathcal{C}_{1}|X^\star\right)|^2}.
    \label{eq:proofeq7}
\end{equation}
Furthermore, considering the establishment of Equation \ref{eq:proofeq6} and Equation \ref{eq:proofeq7}, we state that the definition of $\mathcal{R}_{T^b}(f)$ in Equation \ref{eq:rextra} can be transformed into
\begin{equation}
    \begin{aligned}
        \mathcal{R}_{\mathcal{T}^b}(f) &= \mathcal{R}_{\mathcal{T}^b} + \mathcal{R}_{CLS}\\
        &= \mathcal{R}_{\mathcal{T}^b} + \frac{1}{2}|\nabla P\left(\mathcal{C}_{2}|X^\star\right)-\nabla P\left(\mathcal{C}_{1}|X^\star\right)|\delta_{\Psi_{\mathcal{C}, f,X^\Delta}}^2\\
        &= \mathcal{R}_{\mathcal{T}^b} + \frac{1}{2}|\nabla P\left(\mathcal{C}_{2}|X^\star\right)-\nabla P\left(\mathcal{C}_{1}|X^\star\right)|\cdot \frac{\delta_{\Psi_{\mathcal{C}_1, f,X^A}}^2 + \delta_{\Psi_{\mathcal{C}_2, f,X^A}}^2}{|\nabla P\left(\mathcal{C}_{2}|X^\star\right)-\nabla P\left(\mathcal{C}_{1}|X^\star\right)|^2}\\
        &= \mathcal{R}_{\mathcal{T}^b} + \frac{\delta_{\Psi_{\mathcal{C}_1, f,X^A}}^2 + \delta_{\Psi_{\mathcal{C}_2, f,X^A}}^2}{2|\nabla P\left(\mathcal{C}_{1}|X^\star\right) - \nabla P\left(\mathcal{C}_{2}|X^\star\right)|}\\
        &= \mathcal{R}_{\mathcal{T}^b} + \frac{\delta_{\Psi_{\mathcal{C}_1, f,X^A}}^2 + \delta_{\Psi_{\mathcal{C}_2, f,X^A}}^2}{2|\nabla_{X^A=X^\star} P\left(\mathcal{C}_{1}|X^A\right) - \nabla_{X^A=X^\star} P\left(\mathcal{C}_{2}|X^A\right)|}.
    \end{aligned}
\end{equation}
Concretely, we fit the theorems validated by \cite{tumer1996linear} in our case, and further introduce a required lemma as follows:
\begin{lemma}
\label{lem:3}
Given an independent and sole model $f$ for the bi-classification task $\mathcal{T}^b$, the upper bound of Bayes error can be implemented by
\begin{equation}
    \begin{aligned}
        \mathcal{R}_{\mathcal{T}^b}(f) = \ &\mathcal{R}_{\mathcal{T}^b} + \frac{\delta_{\Psi_{\mathcal{C}_1, f,X^A}}^2 + \delta_{\Psi_{\mathcal{C}_2, f,X^A}}^2}{2|\nabla_{X^A=X^{\star}} P({\mathcal{C}_1}|X^A) - \nabla_{X^A=X^{\star}} P({\mathcal{C}_2}|X^A)|} \\
        \leq \ &\frac{2P({\mathcal{C}_1})P({\mathcal{C}_2})}{1+P({\mathcal{C}_1})P({\mathcal{C}_2})\cdot\left[(\mu_{\mathcal{C}_1}-\mu_{\mathcal{C}_2})^\top\bar{\sigma}^{-1}(\mu_{\mathcal{C}_1}-\mu_{\mathcal{C}_2})\right]} \\
        & + \frac{\delta_{\Psi_{\mathcal{C}_1, f,X^A}}^2 + \delta_{\Psi_{\mathcal{C}_2, f,X^A}}^2}{2|\nabla_{X^A=X^{\star}} P({\mathcal{C}_1}|X^A) - \nabla_{X^A=X^{\star}} P({\mathcal{C}_2}|X^A)|},
    \end{aligned}
\end{equation}
which can be derived by introducing Equation 11, Accordingly, $\delta_{\Psi_{\mathcal{C}_1, f,X^A}}^2$ denotes the variance of the inherent error $\Psi_{\mathcal{C}_1, f,X^A}$, and $\delta_{\Psi_{\mathcal{C}_2, f,X^A}}^2$ denotes the variance of $\Psi_{\mathcal{C}_2, f,X^A}$, respectively.
\end{lemma}

Furthermore, we research on the proposed HTML in an decomposition manner. Following the intuition \cite{DBLP:journals/corr/abs-2109-14433}, HTML can be treated as an implicit multi-loss ensemble learning model \cite{DBLP:journals/corr/abs-2109-14433}, and the features learned by the detached losses are included in different multiple-dimensional subsets of the learned representations. The weights of the classification head can be treated as the implicitly weighting ensemble for representations learned by HTML, such that HTML can be decomposed into the conventional GCL model $f$ and the plug-and-play hierarchical topology isomorphism expertise learning model $f_{\star}$. The combination of the final prediction is defined as follows:
\begin{equation}
    (\left\{f,f_{\star}\right\})(X^A) = \frac{f(X^A) + \gamma \cdot f_{\star}(X^A)}{2},
    \label{eq:proofeq9}
\end{equation}
where $\gamma$ if a coefficient balancing the impacts of the multiple-dimensional subsets of the learned representations that are dependent on the hierarchical topology isomorphism expertise learning model $f_{\star}$, and for the ease of \textit{anaphora resolution}\footnote{The term \textit{anaphora resolution} is used in its idiomatic sense rather than the specific sense in the field of natural language processing \cite{DBLP:conf/iclr/QuSSSC021}.}, we put the graph-tier topology isomorphism expertise learning module and the subgraph-tier topology isomorphism expertise learning module together to form the integrated model $f_{\star}$, such that following Equation 12 in Corollary 5.5, the inherent error of the ensemble model $\left\{f, f_{\star}\right\}$ can be implemented by
\begin{equation}
\begin{aligned}
    \mathcal{E}(\mathcal{C}, \left\{f, f_{\star}\right\}, X^A)&=\Phi_{\mathcal{C}} + \frac{\Psi_{\mathcal{C}, f, X^A}(X^A) + \gamma \cdot \Psi_{\mathcal{C}, f_{\star}, X^A}(X^A)}{2}\\
    &=\Phi_{\mathcal{C}} + \frac{\Psi_{\mathcal{C}, f, X^A}(X^A) + \Psi_{\mathcal{C}, f_{\star}, X^A}(X^A)}{2},
\end{aligned} \label{eq:proofeqmathcaleinherent}
\end{equation}
where we directly resolve $\gamma$ for the ease of deduction, which is achievable because of that $\gamma$ is a constant coefficient that cannot affect the qualitative analysis of the inherent error $\mathcal{E}(\mathcal{C}, \left\{f, f_{\star}\right\}, X^A)$. Concretely, Corollary 5.7 and the corresponding proof in \textbf{Appendix} \ref{app:cor2proof} jointly demonstrate that the only hypothesis base is $\epsilon_{f_{\star}} < \sqrt{3}\epsilon_f$, which is empirically proved, such that the exact value of $\gamma$ does not undermine the establishment of the derived Theorem 5.6 and Corollary 5.7.

According to Equation \ref{eq:proofeq4}, we define the offset ${X^{\Delta}}^\prime$ of the decision boundary for the ensemble model in Equation \ref{eq:proofeq9} as follows:
\begin{equation}
\begin{aligned}
    {X^{\Delta}}^\prime &= \frac{\left(\mathcal{E}\left(\mathcal{C}_{1}, f, X^{A}\right) + \mathcal{E}\left(\mathcal{C}_{1}, f_{\star}, X^{A}\right)\right) -\left(\mathcal{E}\left(\mathcal{C}_{2}, f, X^{A}\right) + \mathcal{E}\left(\mathcal{C}_{2}, f_{\star}, X^{A}\right)\right)}{2\left(\nabla P\left(\mathcal{C}_{2}|X^\star\right)-\nabla P\left(\mathcal{C}_{1}|X^\star\right)\right)} \\
    &= \frac{\mathcal{E}\left(\mathcal{C}_{1}, f, X^{A}\right) + \mathcal{E}\left(\mathcal{C}_{1}, f_{\star}, X^{A}\right) - \mathcal{E}\left(\mathcal{C}_{2}, f, X^{A}\right) + \mathcal{E}\left(\mathcal{C}_{2}, f_{\star}, X^{A}\right)}{2\nabla P\left(\mathcal{C}_{2}|X^\star\right)-2\nabla P\left(\mathcal{C}_{1}|X^\star\right)}.
\end{aligned} \label{eq:proofeqxdeltaprime}
\end{equation}
Considering Equation \ref{eq:rextra} and Equation \ref{eq:proofeq6}, the Bayes error of the ensemble model can be implemented by
\begin{equation}
    \begin{aligned}
        \mathcal{R}_{\mathcal{T}^b}(\left\{f, f_{\star}\right\}) &= \mathcal{R}_{\mathcal{T}^b} + \mathcal{R}_{CLS}^\prime\\
        &= \mathcal{R}_{\mathcal{T}^b} + \frac{1}{2}|\nabla P\left(\mathcal{C}_{2}|X^\star\right)-\nabla P\left(\mathcal{C}_{1}|X^\star\right)|\delta_{\Psi_{\mathcal{C}, f,{X^\Delta}^\prime}}^2.
    \end{aligned}
    \label{eq:proofeq10}
\end{equation}

Thus, we follow the deduction in the aforementioned proof, e.g., Equation \ref{eq:proofeq7}, to derive the variance of ${X^\Delta}^\prime$ as follows:
\begin{equation}
    \delta_{\Psi_{\mathcal{C}, f,{X^\Delta}}^\prime}^2=\frac{\delta_{\Psi_{\mathcal{C}_1, f,X^A}}^2 + \delta_{\Psi_{\mathcal{C}_2, f,X^A}}^2 + \delta_{\Psi_{\mathcal{C}_1, f_{\star},X^A}}^2 + \delta_{\Psi_{\mathcal{C}_2, f_{\star},X^A}}^2}{4|\nabla_{X^A=X^\star} P\left(\mathcal{C}_{1}|X^A\right) - \nabla_{X^A=X^\star} P\left(\mathcal{C}_{2}|X^A\right)|^2}.
    \label{eq:proofeq11}
\end{equation}
By introducing Equation \ref{eq:proofeq11} into Equation \ref{eq:proofeq10}, we can provide the Bayes error bound of the ensemble model:
\begin{equation}
    \mathcal{R}_{\mathcal{T}^b}(\left\{f, f_{\star}\right\})= \mathcal{R}_{\mathcal{T}^b} + \frac{\delta_{\Psi_{\mathcal{C}_1, f,X^A}}^2 + \delta_{\Psi_{\mathcal{C}_2, f,X^A}}^2 + \delta_{\Psi_{\mathcal{C}_1, f_{\star},X^A}}^2 + \delta_{\Psi_{\mathcal{C}_2, f_{\star},X^A}}^2}{8|\nabla_{X^A=X^\star} P\left(\mathcal{C}_{1}|X^A\right) - \nabla_{X^A=X^\star} P\left(\mathcal{C}_{2}|X^A\right)|}.
\end{equation}

In sum, we revisit the deduction and conclusion of Lemma \ref{lem:3}, and derive that the upper bound of the Bayes error is closely associated with the variance of the inherent error, and thus, we can induce the Bayes error upper bound for the proposed method by treating HTML as an implicit ensemble model, such that HTML can be decomposed into the conventional GCL model $f$ and the plug-and-play hierarchical topology isomorphism expertise learning graph model $f_{\star}$. Concretely, we arrive at a summative conclusion based on Lemma \ref{lem:3}, which is detailed by
\begin{corollary}
\label{cor:3}
Given the decomposed HTML models $f$ and $f_{\star}$ for classification, we ensemble the features contained by multiple dimensional subsets of the ultimate representations to generate the prediction in a weighted manner, which is achieved by leveraging the classification weights during linear probing learning, such that the upper bound of Bayes error of the proposed HTML can be approximated by
\begin{equation}
    \begin{aligned}
        \mathcal{R}_{\mathcal{T}^b}(\left\{f, f_{\star}\right\}) = \ &\mathcal{R}_{\mathcal{T}^b} + \frac{\delta_{\Psi_{\mathcal{C}_1, f,X^A}}^2 + \delta_{\Psi_{\mathcal{C}_2, f,X^A}}^2 + \delta_{\Psi_{\mathcal{C}_1, f_{\star}, X^A}}^2 + \delta_{\Psi_{\mathcal{C}_2, f_{\star}, X^A}}^2}{8|\nabla_{X^A=X^{\star}} P({\mathcal{C}_1}|X^A) - \nabla_{X^A=X^{\star}} P({\mathcal{C}_2}|X^A)|} \\
        \leq \ &\frac{2P({\mathcal{C}_1})P({\mathcal{C}_2})}{1+P({\mathcal{C}_1})P({\mathcal{C}_2})\cdot\left[(\mu_{\mathcal{C}_1}-\mu_{\mathcal{C}_2})^\top\bar{\sigma}^{-1}(\mu_{\mathcal{C}_1}-\mu_{\mathcal{C}_2})\right]} \\
        & + \frac{\delta_{\Psi_{\mathcal{C}_1, f,X^A}}^2 + \delta_{\Psi_{\mathcal{C}_2, f,X^A}}^2 + \delta_{\Psi_{\mathcal{C}_1, f_{\star}, X^A}}^2 + \delta_{\Psi_{\mathcal{C}_2, f_{\star}, X^A}}^2}{8|\nabla_{X^A=X^{\star}} P({\mathcal{C}_1}|X^A) - \nabla_{X^A=X^{\star}} P({\mathcal{C}_2}|X^A)|}.
    \end{aligned}
\end{equation}
\end{corollary}
Based on Corollary \ref{cor:3}, we can thus acquire the following proof for Theorem 5.6 as follows:
\begin{proof}
\begin{equation}
\begin{aligned}
    &\left|\mathcal{R}_{\mathcal{T}^b}(f) - \mathcal{R}_{\mathcal{T}^b}(\left\{f, f_{\star}\right\})\right| \\
    = \ &\mathcal{R}_{\mathcal{T}^b}(f) - \mathcal{R}_{\mathcal{T}^b}(\left\{f, f_{\star}\right\}) \\
    = \ &\mathcal{R}_{\mathcal{T}^b} + \frac{\delta_{\Psi_{\mathcal{C}_1, f,X^A}}^2 + \delta_{\Psi_{\mathcal{C}_2, f,X^A}}^2}{2|\nabla_{X^A=X^{\star}} P({\mathcal{C}_1}|X^A) - \nabla_{X^A=X^{\star}} P({\mathcal{C}_2}|X^A)|} \\
    & - \left[\mathcal{R}_{\mathcal{T}^b} + \frac{\delta_{\Psi_{\mathcal{C}_1, f,X^A}}^2 + \delta_{\Psi_{\mathcal{C}_2, f,X^A}}^2 + \delta_{\Psi_{\mathcal{C}_1, f_{\star}, X^A}}^2 + \delta_{\Psi_{\mathcal{C}_2, f_{\star}, X^A}}^2}{8|\nabla_{X^A=X^{\star}} P({\mathcal{C}_1}|X^A) - \nabla_{X^A=X^{\star}} P({\mathcal{C}_2}|X^A)|}\right] \\
    = \ &\frac{\delta_{\Psi_{\mathcal{C}_1, f,X^A}}^2 + \delta_{\Psi_{\mathcal{C}_2, f,X^A}}^2}{2|\nabla_{X^A=X^{\star}} P({\mathcal{C}_1}|X^A) - \nabla_{X^A=X^{\star}} P({\mathcal{C}_2}|X^A)|} \\
    & - \frac{\delta_{\Psi_{\mathcal{C}_1, f,X^A}}^2 + \delta_{\Psi_{\mathcal{C}_2, f,X^A}}^2 + \delta_{\Psi_{\mathcal{C}_1, f_{\star}, X^A}}^2 + \delta_{\Psi_{\mathcal{C}_2, f_{\star}, X^A}}^2}{8|\nabla_{X^A=X^{\star}} P({\mathcal{C}_1}|X^A) - \nabla_{X^A=X^{\star}} P({\mathcal{C}_2}|X^A)|} \\
    = \ & \frac{4\cdot\left(\delta_{\Psi_{\mathcal{C}_1, f,X^A}}^2 + \delta_{\Psi_{\mathcal{C}_2, f,X^A}}^2\right) - \left(\delta_{\Psi_{\mathcal{C}_1, f,X^A}}^2 + \delta_{\Psi_{\mathcal{C}_2, f,X^A}}^2 + \delta_{\Psi_{\mathcal{C}_1, f_{\star}, X^A}}^2 + \delta_{\Psi_{\mathcal{C}_2, f_{\star}, X^A}}^2\right)}{8|\nabla_{X^A=X^{\star}} P({\mathcal{C}_1}|X^A) - \nabla_{X^A=X^{\star}} P({\mathcal{C}_2}|X^A)|} \\
    = \ & \frac{3\cdot\delta_{\Psi_{\mathcal{C}_1, f,X^A}}^2 + 3 \cdot \delta_{\Psi_{\mathcal{C}_2, f,X^A}}^2 - \delta_{\Psi_{\mathcal{C}_1, f_{\star}, X^A}}^2 - \delta_{\Psi_{\mathcal{C}_2, f_{\star}, X^A}}^2}{8|\nabla_{X^A=X^{\star}} P({\mathcal{C}_1}|X^A) - \nabla_{X^A=X^{\star}} P({\mathcal{C}_2}|X^A)|} \\
    \geq \ & 0,
\end{aligned}
\end{equation}
\end{proof}
which can be deduced by adhering the constraint provided in Corollary 5.7.

\subsection{Proof for Corollary 5.7} \label{app:cor2proof}
Initially, we assume that the inherent error, e.g., $\Psi_{\mathcal{C}, f, X^A}$, is bounded by the relatively loose threshold $\epsilon_f$, i.e., $-\epsilon_f \leq \Psi_{\mathcal{C}, f, X^A} \leq \epsilon_f$, where $\epsilon_f$ is a constant. Following Equation \ref{eq:proofeq4} and Equation \ref{eq:proofeq7}, we can provide the corresponding bound of $\delta_{\Psi_{\mathcal{C}, f, X^A}}^2$ by
\begin{equation}
    \begin{aligned}
        \delta_{\Psi_{\mathcal{C}, f, X^A}}^2 &\leq \delta_{\Psi_{\mathcal{C}, f, X^A}}^2 + \mathrm{E}\big[(\epsilon_f-\Psi_{\mathcal{C}, f, X^A})(\Psi_{\mathcal{C}, f, X^A}+\epsilon_f)\big]\\
        &=\epsilon_f^2 - \big(\mathrm{E}[\Psi_{\mathcal{C}, f,X^A}]- \epsilon_f\big)\\
        &\leq \epsilon_f^2,
    \end{aligned}
    \label{eq:proofeq14}
\end{equation}
and thus, we have
\begin{equation}
\left\{
    \begin{aligned}
        \delta_{\Psi_{\mathcal{C}_1, f, X^A}}^2 &\leq \epsilon_f^2\\
        \delta_{\Psi_{\mathcal{C}_2, f, X^A}}^2 &\leq \epsilon_f^2.
    \end{aligned}
\right.
    \label{eq:proofeq15}
\end{equation}
Considering Equation \ref{eq:proofeq14} and Equation \ref{eq:proofeq15}, we can derive
\begin{equation}
    \mathcal{R}_{\mathcal{T}^b}(f) \leq \mathcal{R}_{\mathcal{T}^b} + \frac{\epsilon_f^2}{|\nabla_{X^A=X^\star} P\left(\mathcal{C}_{1}|X^A\right) - \nabla_{X^A=X^\star} P\left(\mathcal{C}_{2}|X^A\right)|}.
    \label{eq:proofeq16}
\end{equation}

Following the aforementioned deduction, we assume that the inherent error $\Psi_{\mathcal{C}, f_{\star}, X^A}$is bounded by a threshold $\epsilon_{f_{\star}}$, i.e., $-\epsilon_{f_{\star}} \leq \Psi_{\mathcal{C}, f_{\star}, X^A} \leq \epsilon_{f_{\star}}$, where $\epsilon_{f_{\star}}$ is constant. Following Equation \ref{eq:proofeqmathcaleinherent} and Equation \ref{eq:proofeqxdeltaprime}, we can provide the upper bound of $\delta_{\Psi_{\mathcal{C}, f, X^A}}^2$ by
\begin{equation}
    \begin{aligned}
        \delta_{\Psi_{\mathcal{C}, f_{\star}, X^A}}^2 &\leq \delta_{\Psi_{\mathcal{C}, f_{\star}, X^A}}^2 + \mathrm{E}\big[(\epsilon_{f_{\star}}-\Psi_{\mathcal{C}, {f_{\star}}, X^A})(\Psi_{\mathcal{C}, {f_{\star}}, X^A}+\epsilon_{f_{\star}})\big]\\
        &=\epsilon_{f_{\star}}^2 - \big(\mathrm{E}[\Psi_{\mathcal{C}, {f_{\star}},X^A}]- \epsilon_{f_{\star}}\big)\\
        &\leq \epsilon_{f_{\star}}^2,
    \end{aligned}
    \label{eq:proofeq17}
\end{equation}
and thus, we have
\begin{equation}
\left\{
    \begin{aligned}
        \delta_{\Psi_{\mathcal{C}_1, {f_{\star}}, X^A}}^2 &\leq \epsilon_{f_{\star}}^2\\
        \delta_{\Psi_{\mathcal{C}_2, {f_{\star}}, X^A}}^2 &\leq \epsilon_{f_{\star}}^2.
    \end{aligned}
\right.
    \label{eq:proofeq18}
\end{equation}
Thus, we follow Equation \ref{eq:proofeq17} and Equation \ref{eq:proofeq18} to deduce that
\begin{equation}
    \mathcal{R}_{\mathcal{T}^b}(\left\{f, f_{\star}\right\}) \leq \mathcal{R}_{\mathcal{T}^b} + \frac{\epsilon_f^2 + \epsilon_{f_{\star}}^2}{4|\nabla_{X^A=X^\star} P\left(\mathcal{C}_{1}|X^A\right) - \nabla_{X^A=X^\star} P\left(\mathcal{C}_{2}|X^A\right)|}.
    \label{eq:proofeq19}
\end{equation}

Concretely, if the only constraint, i.e., $\epsilon_{f_{\star}} < \sqrt{3} \cdot \epsilon_f$, holds, considering Equation \ref{eq:proofeq16} we can transform Equation \ref{eq:proofeq19} into the following conclusive statement that
\begin{equation}
\begin{aligned}
    \mathcal{R}_{\mathcal{T}^b}(\left\{f, f_{\star}\right\}) &\leq \mathcal{R}_{\mathcal{T}^b} + \frac{\epsilon_f^2 + \epsilon_{f_{\star}}^2}{4|\nabla_{X^A=X^\star} P\left(\mathcal{C}_{1}|X^A\right) - \nabla_{X^A=X^\star} P\left(\mathcal{C}_{2}|X^A\right)|}\\
    &< \mathcal{R}_{\mathcal{T}^b} + \frac{\epsilon_f^2}{|\nabla_{X^A=X^\star} P\left(\mathcal{C}_{1}|X^A\right) - \nabla_{X^A=X^\star} P\left(\mathcal{C}_{2}|X^A\right)|}\\
    &= O\left(\mathcal{R}_{\mathcal{T}^b}(f)\right),
\end{aligned} \label{eq:proofeq20}
\end{equation}
which holds due to the establishment of
\begin{equation}
\begin{aligned}
    & \frac{\epsilon_f^2 + \epsilon_{f_{\star}}^2}{4} - \epsilon_f^2\\
    = &\frac{\epsilon_f^2 + \epsilon_{f_{\star}^2} - 4 \cdot \epsilon_f^2}{4}\\
    = &\frac{\epsilon_{f_{\star}^2} - 3 \cdot \epsilon_f^2}{4}\\
    = &\frac{\left(\epsilon_{f_{\star}} + \sqrt{3} \cdot \epsilon_f^2\right)\left(\epsilon_{f_{\star}} - \sqrt{3} \cdot \epsilon_f^2\right)}{4}\\
    < & 0.
\end{aligned} \label{eq:proofeq21}
\end{equation}
Therefore, the upper bound of $\mathcal{R}_{\mathcal{T}^b}(\left\{f, f_{\star}\right\})$ is relatively tighter than that of $\mathcal{R}_{\mathcal{T}^b}(f)$, such that compared with the independent conventional GCL model, the proposed HTML acquires a tighter Bayes error upper bound, thereby proving the establishment of Corollary 5.7.

\section{Datasets, Baselines and Training in Various Settings}\label{app:data}

\subsection{Unsupervised Representation Learning}\label{C:1}
\textbf{Datasets}.
We present the statistics of the datasets utilized for unsupervised learning in Table \ref{tab:data1}.
\begin{table}[ht]
\caption{Datasets statistics for unsupervised learning.}
\label{tab:data1}
\centering
\begin{tabular}{ccccc}
\hline
 Datasets & Category & Graph Num. & Avg. Node & Avg. Degree \\
\hline
 NCI1 & Biochemical Molecules & 4110 & $29.87$ & $1.08$ \\
PROTEINS & Biochemical Molecules & 1113 & $39.06$ & $1.86$ \\
DD & Biochemical Molecules & 1178 & $284.32$ & $2.51$ \\
MUTAG & Biochemical Molecules & 188 & $17.93$ & $1.10$ \\
\hline
COLLAB & Social Networks & 5000 & $74.49$ & $32.99$ \\
RDT-B & Social Networks & 2000 & $429.63$ & $1.15$ \\
RDB-M & Social Networks & 4999 & $508.52$ & $1.16$ \\
IMDB-B & Social Networks & 1000 & $19.77$ & $4.88$ \\
\hline
\end{tabular}
\end{table}
\\
\textbf{Baselines.} To validate the effectiveness of HTML with unsupervised representation learning setting, we compare it with the following state-of-the-art baselines: (1) SOTA graph kernel methods such as graphlet kernel (GL) \cite{DBLP:journals/jmlr/ShervashidzeVPMB09}, Weisfeiler-Lehman sub-tree kernel (WL) \cite{DBLP:journals/jmlr/ShervashidzeSLMB11}and deep graph kernel (DGK) \cite{DBLP:conf/kdd/YanardagV15}; (2) graph-level representation learning method InfoGraph \cite{DBLP:conf/iclr/SunHV020}; (3) graph contrastive learning methods such as GraphCL \cite{DBLP:conf/nips/YouCSCWS20}, joint augmentation
optimization (JOAO) \cite{DBLP:conf/icml/YouCSW21}, AD-GCL \cite{suresh2021adversarial}, rationale-aware graph contrastive
learning (RGCL) \cite{DBLP:conf/icml/LiWZW0C22}, and prototypical graph contrastive learning (PGCL) \cite{DBLP:journals/corr/abs-2106-09645}.

\textbf{Implementation.} GraphCL+HTML, PGCL+HTML, and RGCL+HTML are implemented with GraphCL, PGCL, and RGCL as the backbone, respectively. Following the same training principles as the benchmark methods, GraphCL+HTML and PGCL+HTML are performed five times with different seeds, while RGCL+HTML is performed ten times, with the mean and standard deviation of accuracies (\%) reported. The hyper-parameters $\alpha$  and $\beta$ are tuned in $\{1,1\times10^1,1\times10^2,1\times10^3,1\times10^4\}$ respectively.

\subsection{Transfer Learning}
\textbf{Datasets}.
We present the statistics of the datasets utilized for transfer learning in Table \ref{tab:data2}. 
\begin{table}[ht]
\caption{Datasets statistics for transfer learning.}
\label{tab:data2}
\centering
\begin{tabular}{cccccc}
\hline
Datasets & Category & Utilization & Graph Num. & Avg. Node & Avg. Degree \\
\hline
ZINC-2M & Biochemical Molecules & Pre-Training & 2000000 & $26.62$ & $57.72$ \\
\hline
BBBP & Biochemical Molecules & Finetuning & 2039 & $24.06$ & $51.90$ \\
Tox21 & Biochemical Molecules & Finetuning & 7831 & $18.57$ & $38.58$ \\
ToxCast & Biochemical Molecules & Finetuning & 8576 & $18.78$ & $38.52$ \\
SIDER & Biochemical Molecules & Finetuning & 1427 & $33.64$ & $70.71$ \\
ClinTox & Biochemical Molecules & Finetuning & 1477 & $26.15$ & $55.76$ \\
MUV & Biochemical Molecules & Finetuning & 93087 & $24.23$ & $52.55$ \\
HIV & Biochemical Molecules & Finetuning & 41127 & $25.51$ & $54.93$ \\
BACE & Biochemical Molecules & Finetuning & 1513 & $34.08$ & $73.71$ \\
\hline
\end{tabular}
\end{table}
\\
\textbf{Baselines.} For the transfer learning setting, we compare HTML with the following benchmark methods: graph-level unsupervised learning method Infomax \cite{DBLP:conf/iclr/VelickovicFHLBH19}; various pre-training methods for graph neural networks such as EdgePred \cite{DBLP:conf/iclr/HuLGZLPL20}, AttrMasking \cite{DBLP:conf/iclr/HuLGZLPL20}, and ContextPred \cite{DBLP:conf/iclr/HuLGZLPL20}; diverse graph contrastive learning methods such as GraphCL \cite{DBLP:conf/nips/YouCSCWS20}, JOAO \cite{DBLP:conf/icml/YouCSW21}, AD-GCL \cite{suresh2021adversarial}, and RGCL \cite{DBLP:conf/icml/LiWZW0C22}.

\textbf{Implementation}. To conduct experiments with transfer learning setting, we pre-trained the ZINC-2M dataset under the best hyper-parameters and fine-tuned it ten times on eight biochemical molecule datasets. The hyper-parameters $\alpha$ and $\beta$ are chosen in the same range as \ref{C:1}.

\subsection{Large-scale Graph Representation Learning}
\textbf{Datasets.} We exhibit the details of the dataset used for large-scale graph representation learning in Table \ref{tab:data3}.

\begin{table}[ht]
\caption{Datasets statistics for large-scale graph representation learning.}
\label{tab:data3}
\centering
\begin{tabular}{ccccc}
\hline 
Datasets &  Nodes &Edges& Features& Classes\\
\hline
Amazon Photos & 7,650 &  119,081 & 745 &  8 \\
\hline
\end{tabular}
\end{table}

\textbf{Baselines.} We employed the following baselines for comparison with our method: MVGRL \cite{DBLP:conf/icml/HassaniA20}, a graph contrastive learning method based on structural view augmentation; graph-level unsupervised learning method Random-Init \cite{DBLP:conf/iclr/VelickovicFHLBH19}; node-level graph contrastive learning method GRACE \cite{DBLP:journals/corr/abs-2006-04131}; scalable graph self-supervised representation learning method BGRL \cite{DBLP:conf/iclr/ThakoorTAADMVV22} for very large graphs.

\textbf{Implementation.} We implement HTML with BGRL as the backbone and report the results of training 10,000 epochs. The hyper-parameters $\alpha$ and $\beta$ are chosen in the same range as \ref{C:1}.
\section{Illustrative Example} \label{app:example}
\begin{figure*}[ht]
\centering
\includegraphics[scale=0.7]{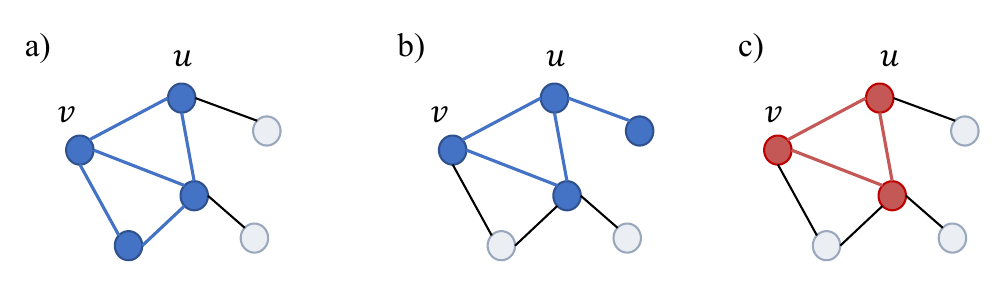}
\vskip -0.1in
\caption{An example of an overlap subgraph. a) and b) demonstrate the neighborhood subgraph of $v$ and $u$ with dark \textcolor{blue}{blue} circles and lines, respectively, and c) represents the overlap subgraph of $v$ and $u$ with \textcolor{red}{red} circles and lines.}
\label{fig:overlap}
\vskip -0.1in
\end{figure*}

Figure \ref{fig:overlap} shows an example of an overlap subgraph.


\end{document}